\documentclass[10pt,twocolumn,letterpaper]{article}

\usepackage[pagenumbers]{cvpr} % To force page numbers, e.g. for an arXiv version

\usepackage{graphicx}
\usepackage{amsmath}
\usepackage{amssymb}
\usepackage{booktabs}

\usepackage{epstopdf}
\graphicspath{{../}}

\usepackage[pagebackref,breaklinks,colorlinks]{hyperref}

\usepackage[capitalize]{cleveref}
\crefname{section}{Sec.}{Secs.}
\Crefname{section}{Section}{Sections}
\Crefname{table}{Table}{Tables}
\crefname{table}{Tab.}{Tabs.}

\def\cvprPaperID{8677} % *** Enter the CVPR Paper ID here
\def\confName{CVPR}
\def\confYear{2022}

\usepackage{multirow} %% For Errors table
\usepackage{microtype}
\makeatletter
\@namedef{ver@everyshi.sty}{}
\makeatother

\usepackage
[disable]
{todonotes}

\definecolor{gold}{rgb}{1.0, 0.55, 0.0}
\definecolor{forrestgreen}{rgb}{0.1, 0.4, 0.1}

\definecolor{darkorange}{rgb}{0.8, 0.55, 0.0}

\usepackage{pgfplots}
\newcommand{\subtitle}[1]{
    \newpage
    \null
    \begin{center}
        \iftoggle{cvprrebuttal}{{\large \bf #1 \par}}{{\Large \bf #1 \par}}
        \iftoggle{cvprrebuttal}{\vspace*{-22pt}}{\vspace*{24pt}}
    \end{center}
}

\begin{document}

%%%%%%%%% TITLE - PLEASE UPDATE
\title{N-SfC: Robust and Fast Shape Estimation from Caustic Images} 
% \vspace{-.5em}}
%
% \author{First Author\\
% Institution1\\
% Institution1 address\\
% {\tt\small firstauthor@i1.org}
% % For a paper whose authors are all at the same institution,
% % omit the following lines up until the closing ``}''.
% % Additional authors and addresses can be added with ``\and'',
% % just like the second author.
% % To save space, use either the email address or home page, not both
% \and
% Second Author\\
% Institution2\\
% First line of institution2 address\\
% {\tt\small secondauthor@i2.org}
% }
% \maketitle
\author{
\hspace{1.6em}Marc Kassubeck$^1$\hspace{1.8em}
Moritz Kappel$^1$\hspace{1.8em}
Susana Castillo$^1$\hspace{2.6em}
Marcus Magnor$^1$\vspace{0.4em}\\
% Affiliations
{\parbox{\textwidth}{\centering \small $^1$ Computer Graphics Lab, TU Braunschweig, Germany
%\\
% {\tt\small  \{kassubeck,kappel,castillo,magnor\}@cg.cs.tu-bs.de}\\
\hspace{7pt}{\tt\small \{lastName\}@graphics.tu-bs.de}
      }
    }
}
\twocolumn[{ 
\renewcommand\twocolumn[1][]{#1} 
\maketitle 
\begin{center} 
    \vspace{-2em}
    \includegraphics[width=\textwidth, keepaspectratio]{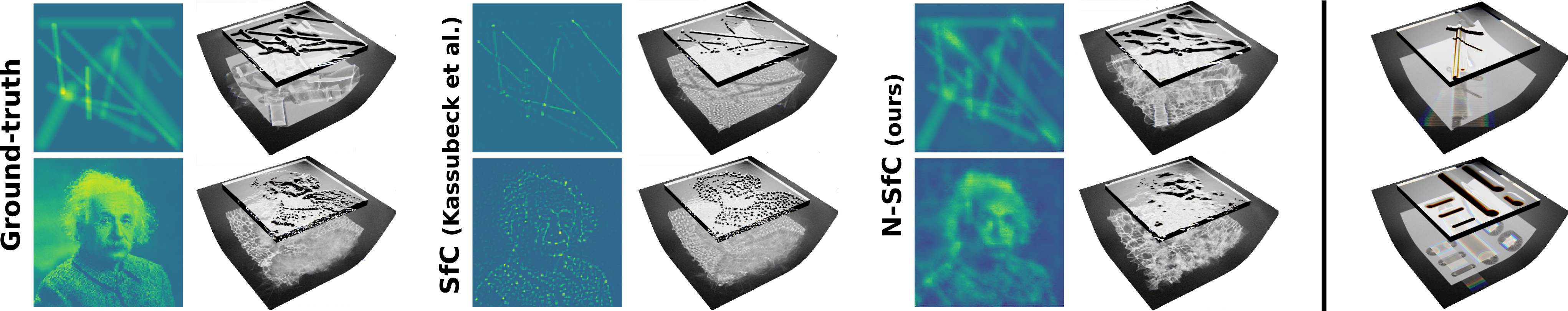}
    \captionof{figure}{
    \textbf{Our Neural - Shape from Caustics (N-SfC) framework} estimates the shape of a glass substrate from a single observation of the resulting caustic image \textit{(far left)}. 
    We extend existing work based on physically based light transport simulation \textit{(middle left)} with learned denoising and gradient descent \textit{(middle right)}. Multispectral simulation and dispersion effects are shown on the \textit{far right}.
    }
    \label{fig:teaser} 
\end{center} 
}] 
%
%%%%%%%%%%%%%%%%%%%%%%%%%%%%%%%%%%%%%%%%%%%%%%%%%%%%%%%%%%%%%%%%%%%%%%%%%%%%%%

%%%%%%%%% ABSTRACT
%%%%%%%%% ABSTRACT
\begin{abstract}
    This paper deals with the highly challenging problem of reconstructing the shape of a refracting object from a single image of its resulting caustic. %, i.e. the brightness distribution on a surface caused by illuminating the object.
    Due to the ubiquity of transparent refracting objects in everyday life, reconstruction of their shape entails a multitude of practical applications.
    The recent Shape from Caustics (SfC) method casts the problem as the inverse of a light propagation simulation for synthesis of the caustic image, that can be solved by a differentiable renderer.
    However, the inherent complexity of light transport through refracting surfaces currently limits the practicability with respect to reconstruction speed and robustness.
    To address these issues, we introduce Neural-Shape from Caustics (N-SfC), a learning-based extension that incorporates two components into the reconstruction pipeline: 
    a denoising module, which alleviates the computational cost of the light transport simulation, and an optimization process based on learned gradient descent, which enables better convergence using fewer iterations.
    Extensive experiments demonstrate the effectiveness of our neural extensions in the scenario of quality control in 3D glass printing, where we significantly outperform the current state-of-the-art in terms of computational speed and final surface error.
\end{abstract}
    % This paper deals with the highly challenging problem of reconstructing the shape of a refracting object from a single image of its resulting caustic. Due to the ubiquity of transparent refracting objects in everyday life, reconstruction of their shape entails a multitude of practical applications. The recent Shape from Caustics (SfC) method casts the problem as the inverse of a light propagation simulation for synthesis of the caustic image, that can be solved by a differentiable renderer. However, the inherent complexity of light transport through refracting surfaces currently limits the practicability with respect to reconstruction speed and robustness. To address these issues, we introduce Neural-Shape from Caustics (N-SfC), a learning-based extension that incorporates two components into the reconstruction pipeline: a denoising module, which alleviates the computational cost of the light transport simulation, and an optimization process based on learned gradient descent, which enables better convergence using fewer iterations. Extensive experiments demonstrate the effectiveness of our neural extensions in the scenario of quality control in 3D glass printing, where we significantly outperform the current state-of-the-art in terms of computational speed and final surface error. 
\vspace{-1em}
%
%%%%%%%%%%%%%%%%%%%%%%%%%%%%%%%%%%%%%%%%%%%%%%%%%%%%%%%%%%%%%%%%%%%%%%%%%%%%%%
%%%%%%%%% BODY TEXT
\section{Introduction}\label{sec:intro}
Recent advances in physics-based differentiable rendering have enabled to incorporate ever more complex light transport effects such as caustics into inverse vision problems. 
Previously, these effects had to be modeled by hand in a time consuming fashion, while readily available gradients from differentiable rendering frameworks allow application-specific reconstructions to focus more on efficient regularization and general optimization schemes.
One of these application-specific examples is the Shape from Caustics (SfC) problem as formulated by Kassubeck~\etal~\cite{kassubeck2021}, which deals with the under-constrained problem of reconstructing the shape of a refracting object from its resulting caustic image, \ie the brightness distribution as seen on a screen surface under illumination from a known light source.
This problem has many applications in inline quality control of optical components, especially in the emerging field of integrated optical manufacturing as enabled by \eg the additive fabrication or 3D printing of glass.
The requirements on the feedback and quality control loop are such that it is desirable to take the measurement \textit{in-situ}, such as after each printed layer, and being able to take the reconstruction result into account for the next layer deposition to compensate for small deviations in previous layers, excluding established high precision but time consuming measurements such as surface profilometry and confocal microscopy.
Thus, the inline measurement is not meant to replace established high accuracy measurement methods, but complement it in the most time-sensitive cases.
Furthermore, with the trend in fabrication to produce smaller and smaller batch sizes down to batch sizes of one in \textit{individualized manufacturing}, the control system needs to robustly assess a large variety of possible shapes, which are not known a priori.
To address these problems, SfC~\cite{kassubeck2021} proposed an optical measurement setup with no moving parts, which requires only a single caustic image as the input to their classical constrained optimization based reconstruction.
However, when trying to adapt their method to the real production process, the reconstruction algorithm revealed deficits in terms of reconstruction speed and robustness.

We address these problems by introducing two learned components into the reconstruction loop: a denoiser and a learned gradient descent scheme.
Since the image formation simulation is based on Monte-Carlo integration and unless a large amount of samples are taken into account, the resulting caustics are invariably noisy, hindering the forward simulation and backpropagation.
Our denoiser alleviates this problem by allowing us to fix the number of forward simulation samples, keeping the runtime cost in check.
Secondly, our learned gradient descent scheme allows to leave out a significant amount of gradient descent steps, while still producing results with lower total shape error.
We show empirically that this combination allows to avoid spurious local minima, enabling more robust reconstructions.

\iftoggle{cvprfinal}
{Our full code including dataset generation scripts can be found here: \small\url{https://graphics.tu-bs.de/publications/kassubeck2021n-sfc}}
{We will provide the full code including dataset generation scripts upon publication of this paper.} 
\section{Related Work}\label{sec:rw} 
As our work relates to differentiable and inverse rendering as well as to computational caustic design, denoising, learned gradient descent and refractive reconstruction methods, we will group the related works into these categories:

\textit{Differentiable rendering} describes a subset of the field of inverse rendering, \ie estimating physically valid parameters from imaging data, by means of a differentiable renderer.
Those renderers formulate the forward image formation process in a differentiable manner and allow to efficiently compute gradients with respect to relevant free parameters, which can in turn be used in gradient-based local parameter search strategies.
Redner~\cite{li2018}, Mitsuba 2~\cite{nimier2019, loubet2019, nimier2020} and Path replay backpropagation~\cite{vicini2021} are increasingly efficient and fully featured renderers but in their current iteration mainly use unidirectional path tracing for image formation and are thus not applicable to our caustics-based optimization problem without further adaptations.
Several other authors~\cite{li2018, loubet2019, bangaru2020, zhang2020, zhang2021} all tackle the difficult problem of providing gradients with respect to scene geometry, which is in the mesh case not trivially differentiable due to visibility discontinuities.
Unbiased gradient estimators~\cite{bangaru2020, zhang2020} have been shown to be beneficial in terms of convergence behavior and final parameter estimation results~\cite{luan2021}, but for our problem even perfect gradients would only achieve convergence to an improper local minimum in many cases. 
Thus we focus our efforts on a learned gradient descent system to circumvent those spurious local minima an achieve a plausible reconstruction for the 3D printing process at hand.
We follow the simulation approach from SfC~\cite{kassubeck2021} for image formation and gradient generation, as they in turn build upon~\cite{frisvad2014}, which allows for sharper caustic edges with fewer samples.
% A recent work~\cite{lyu2020} reports estimation of shape of general glass objects with exactly two refractions with a differentiable rendering system utilizing multiple views and structured light. 
% This approach achieves high quality results of free form shapes, but the physical setup complexity is beyond integration in existing manufacturing machines and the scope of this paper.

\textit{Caustic design} is stated as the problem of finding a refractor or reflector, which produces a desired caustic image on a given screen surface.
Schwartzburg~\etal~\cite{schwartzburg2014} define caustic design as a two step process by first solving an \textit{optimal transport problem} and then calculating a height field achieving said transport.
Meyron~\etal~\cite{meyron2018} expand upon the theory of optimal transport and provide a general algorithm for different computational design tasks.
% However as far as physical constraints on the shape are concerned, those methods mostly enforce smoothness and reduced curvature to ensure physical realisability, whereas we are concerned with reconstructing true shapes given data of achievable geometries.
However, regarding physical constraints on the shape, those methods mostly enforce smoothness and reduced curvature to ensure physical realisability, whereas we are concerned with reconstructing true shapes given data of achievable geometries.

\textit{Denoising} plays an important role in many physics-based image formation simulations~\cite{huo2021}, as the Monte-Carlo nature of path-tracing-based methods quickly leads to correct but highly noisy results, which clear up at the rate of the square root of the number of samples.
Thus filtering methods to alleviate the strain on computational resources are a component of even commercial rendering systems~\cite{arnold2021}.
With the rise of neural-network-based image processing, state-of-the-art denoising methods often integrate learned components into the processing pipeline.
One can categorize these methods, based on whether they operate on the final output image~\cite{kalantari2015, bako2017, vogels2018, chaitanya2017} or act deeper in the path tracing process to predict global illumination effects~\cite{nalbach2017} high-resolution radiance maps from low resolution samples~\cite{jiang2021}.
Our method directly operates in image-space, however we note that noise statistics of photon mapping methods differs from regular unidirectional path-tracing due to the bias in the estimator~\cite{zeng2020, zhu2020}. 
In this case regular pre-trained denoiser is expected to perform sub-par, thus we opt to create our own dataset and network.
By design of our denoiser performs a similar task to learned density estimation~\cite{zhu2020} as part of photon mapping based image generation pipeline.
However, we further incorporate this denoiser into an inverse parameter estimation problem, which has not been presented before to the best of our knowledge.

\textit{Learned gradient descent} as a subset of \textit{meta-learning} or \textit{learning to learn} replaces the classical hand-designed optimizer with a learned component~\cite{Schmidhuber1992,Schmidhuber1993,Schmidhuber1997,Younger2001,Hochreiter2001}, improving ill-posed inverse problems by including a prior on reachable solutions by modification of gradients.
Additionally, larger parameter-specific steps can be taken, allowing for faster convergence.
We take inspiration from recent work~\cite{andrychowicz2016, flynn2019}, which cast the optimization trajectory as steps in a recurrent neural network, conditioned on (approximate) gradients and other problem specific inputs.

Several works tackle the \textit{reconstruction of refractive shapes} by proposing the inclusion of other sources of information such as multiple views and checkerboard patterns~\cite{morris2005} or color coded light fields~\cite{Wetzstein2011}. 
For example, a recent work~\cite{lyu2020} reports estimation of shape of general glass objects with a differentiable rendering system utilizing multiple views and Gray coded structured light. 
This approach achieves high quality results of free form shapes, but the physical setup complexity is beyond integration in existing manufacturing machines and the scope of this paper.
We restrict ourselves to a single view and an arbitrary light source, but create a large dataset to build a prior over achievable shapes.
In the preparation of this dataset we are related to but more specialized than~\cite{SuperCaustics}, who provide a general dataset for computer vision tasks for scenes with transparent refracting objects.
 
\section{Method}\label{sec:method}
\begin{figure}
    \centering
    \includegraphics[width=\linewidth, keepaspectratio]{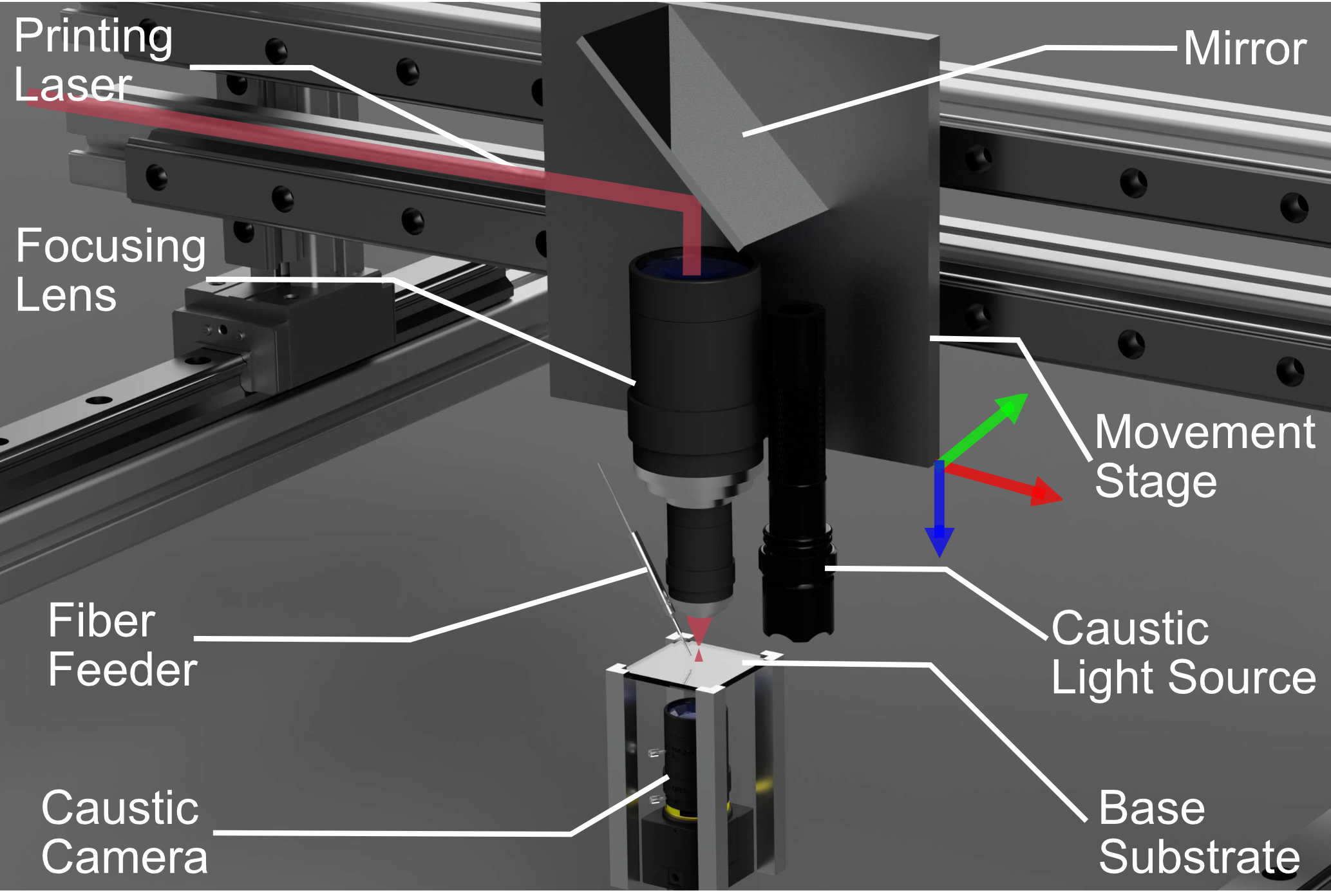}
\caption{\textbf{Schematic sketch of \textit{in-situ} feedback during production}:
compared to the usual setup~\cite{Von-Witzendorff2018}, this includes only two additional components:
    a light source above and a camera below the substrate to generate and capture caustic images.}
    \label{fig:capture_sketch}
\end{figure}

Following the approach of SfC~\cite{kassubeck2021} we represent the solution space of our shape reconstruction as a height field $h \in \mathbf{R}^{n\times n}$ over the flat base substrate of known thickness $d$.
To simulate the resulting caustic image we place a light source above the substrate and calculate the wavelength-dependent irradiance $E \in \mathbf{R}_{+}^{n_w \times m\times m}$ per pixel of a sensor surface below the substrate.
The simulation (cf.~\cite{frisvad2014}) and subsequent gradient calculation through backpropagation mainly depends on three hyperparameters: The number of samples $n_l$, \ie the number of paths outgoing from the light, the number of distinct wavelengths associated with each ray $n_w$ and a smoothing parameter, which controls the size of an elliptical footprint over which the energy of each transported photon is distributed.
A more in-depth overview of the parameters is given in our supplemental material.

To illustrate a potential real-world application of this setup we provide a sketch in~\cref{fig:capture_sketch}, where this setup is integrated into a manufacturing process, which concerns glass 3D printing by selectively melting glass fibers onto a base substrate.
Note that the only moving parts in this are the ones, which were already present in the production process, we simply added the light source above the substrate and a camera with appropriate field of view and focus distance below.
While having the advantage of being easily integrable into even existing setups, this also leads to added complexity for the reconstruction process as we discuss in \cref{ssec:underdeterminism}.
We note that this production example informs physical measurements and dataset distribution as discussed in \cref{ssec:data}, but the methodology of \cref{ssec:pipeline} is independent of this specific application case. 
The methodology can help in similar scenarios, where camera and light placement as well as the compute budget restrict the amount of data that can be obtained or processed. 

\subsection{Underdeterminism of the Problem}\label{ssec:underdeterminism}
%
%%      FIGURE UNDERDETERMINISM
\begin{figure}[tb]
    \centering
    \includegraphics[width=\linewidth]{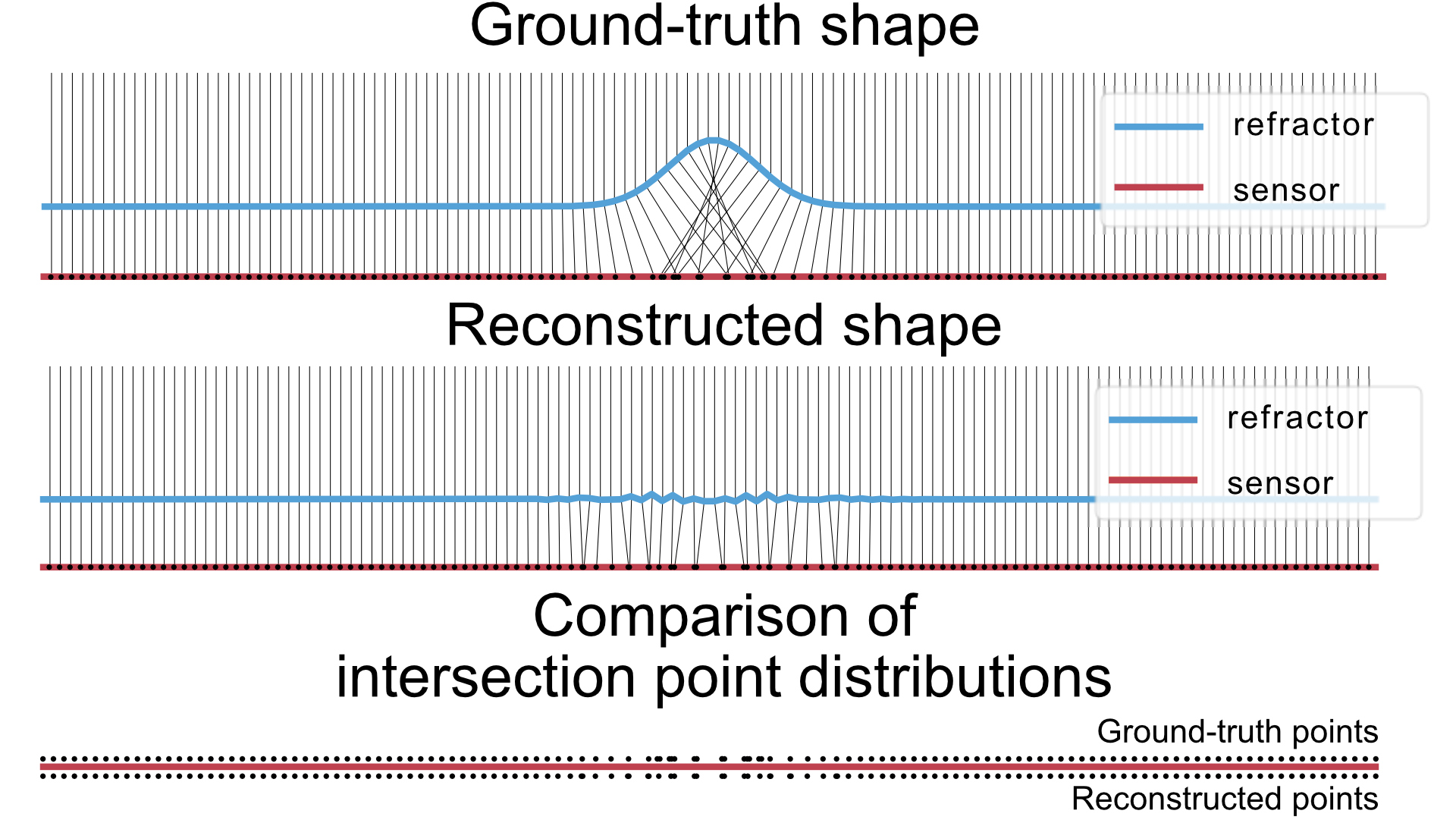}
\caption{\textbf{Underdeterminism of the Problem}: a given refracting shape (top) is reconstructed by optimizing for the Hausdorff distance of intersection points (middle). The resulting shape, albeit near perfect in the metric considered (bottom) is still far from the desired solution, illustrating ill-posedness of such problems.}
    \label{fig:underdeterminism}
\end{figure}
%
%%      FIGURE PIPELINE
\begin{figure*}[th]
    \centering
    \includegraphics[width=\textwidth, keepaspectratio]{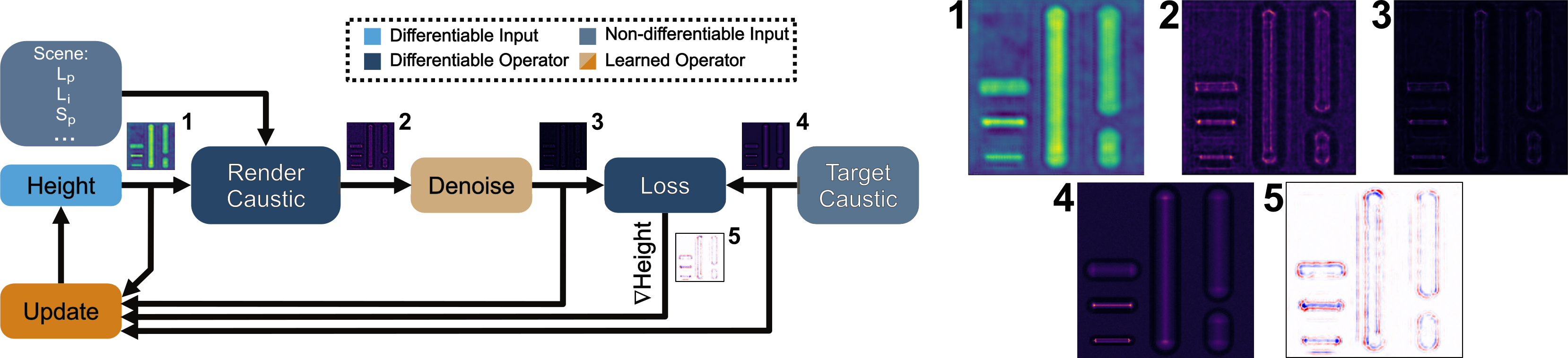}
    \caption{\textbf{Our processing pipeline}: it includes differentiable building blocks and trained components. 
    \texttt{Height}, \texttt{Scene} and \texttt{Target Caustic} are inputs into the method and the updated \texttt{Height} is the final output.
    1-5 display representative outputs of each step.
    1 and 5 are scalar valued and 2-4 are vector valued images with $n_w$ channels.
    This case displays the $L_2$-norm over the channel dimension.
    We use the same color map for data of equivalent type throughout the paper.
    }
    \label{fig:pipeline}
\end{figure*}
%%%
We show that reconstruction of a refracting shape from a single caustic image is severely under-constrained by considering a simple 2D toy example.
\cref{fig:underdeterminism} shows a ground-truth shape and resulting refracted light paths using a fixed refractive index as well as a reconstruction using the Hausdorff metric between the point sets of ground truth and estimated intersection.
This can be thought of as roughly equivalent to the irradiance-based metric in \cref{ssec:pipeline}, when each light path transports the same amount of energy.
The result is obtained by optimizing the height of the refracting shape with the (biased) gradients of SfC~\cite{kassubeck2021} and with the Adam optimizer~\cite{kingma2017adam} of PyTorch~\cite{paszke2017, paszke2019}. 
The initial guess was set to a flat surface.
When comparing the intersection distributions it becomes clear that for this intersection plane the estimation has reached near perfect parity, even with only very minor and local shape changes.
Note that this problem would also persist when using unbiased gradients which is in contrast to other recent work~\cite{luan2021}, which reported convergence to better minima when using unbiased estimators.
This is because even the biased estimator finds a local minimum that is close to a true global minimum in the given loss function.

Classical approaches to deal with this problem are to include more data about the true light paths into the problem.
This could be achieved by obtaining the intersection points (or equivalently caustic images) at different depths, coding the light paths spatially by projecting different colors~\cite{Wetzstein2011} or temporally by projecting varying patterns~\cite{morris2005}. 
However in our case we wish to consider the most general case, where only a single image under a given light source can be obtained.

Given those constraints the modelling has to include a prior on the solution space to disambiguate solutions. 
The prior can be hand-crafted~\cite{kassubeck2021} or learned from data as in this work.
This is motivated by the fact that the generating process we wish to consider here is 3D printing by laser glass deposition~\cite{Von-Witzendorff2018}, which invalidates very small rapidly changing structures like in the reconstruction of \cref{fig:underdeterminism}.
Furthermore the integration into a manufacturing process allows continuous improvement of the prior by adding newly manufactured and measured samples into such a dataset, making it an ideal candidate for such modeling.
\subsection{Pipeline Overview}\label{ssec:pipeline}
\cref{fig:pipeline} shows an overview of the processing steps of the proposed method.
Starting from an initial guess of the refractor height field and other non-differentiable scene parameters we compute the caustic with a differentiable rendering module~\cite{kassubeck2021} and pass the result through a learned denoising module.
This caustic image is then compared to the desired target caustic and after backpropagating the gradient to the initial height field this gradient along with the initial guess, the denoised caustic image and the target caustic image is then fed into an update module, which computes the adjustment of the height field.
The whole process is then potentially looped until a pre-defined convergence criterium is met.

To be more specific the non-differentiable scene parameters define the rest of the scene and potential objects, which interact with the light paths.
Summarizing these parameters as a vector $\theta \in \mathbb{R}^{n_\theta}$, our rendering function can be defined as
\begin{equation}
    R : \mathbb{R}^{n \times n} \times \mathbb{R}^{n_\theta} \rightarrow \mathbb{R}_{+}^{n_w \times m \times m},
\end{equation}
where $n \times n$ and $m \times m$ indicate the pixel resolution of the height map and the caustic image respectively, and $n_w$ denotes the number of wavelengths in the simulation.
Its output is the wavelength-dependent irradiance $E$ at the sensor plane.
Note that unlike many other image processing tasks we cannot impose an a priori upper limit on this quantity, since it is largely dependent on the intensity of the light source and focus due to the estimated geometry.
Consequently all subsequent steps need to handle these full dynamic range images.
The first of these is the denoising network $D$, which is an \textit{endofunction}:
\begin{equation}
    D : \mathbb{R}_{+}^{n_w \times m \times m} \rightarrow \mathbb{R}_{+}^{n_w \times m \times m}.
\end{equation}
The intuition behind this component in the context of inverse problems is that usually computing and memory budget is limited when trying to design a fast feedback algorithm. 
This directly leads to limiting the number of samples $n_l$ in image and gradient computation.
However the target caustic is of a different distribution, because it is either obtained by direct measurement or is a high quality simulation that is largely devoid of noise.
Recent work~\cite{CVPRTut2021} suggests that unbiased noise in gradient descent optimization is not the limiting factor, when combined with appropriate optimizers, but note that in this case we have additional bias in the simulation.
Thus we make an effort to transform the estimated caustic image into the same distribution as the target caustics under consideration with the denoising component.
An important observation is that the ajoint of the denoising operator degrades the gradient signal coming from the loss function.
Thus we exclude the denoiser from the backward pass and replace it with an identity function in this case.

\subsection{Network Architecture}\label{ssec:architecture}
Subsequent processing of caustic images is dependent on two neural-network components, which belong to the same architectural family (see our supplemental material).
Both networks share a structure similar to UNet~\cite{UNet2015} as the main component and differ in a few blocks with respect to the input and output. 
The denoiser includes a single \texttt{Conv + nonlin} block, which expands the number of channels to $c_{init}\in [1, 32]$ channels for the UNet part of the network.
An equivalent block contracts those channels after the UNet part of the denoising network.

The loss function we employ in \cref{fig:pipeline} is the mean squared error of irradiances, so the full objective function is
\begin{equation}
    \mathbf{L_{irrad}}(h, \hat{E}) = \frac{1}{n_w m^2} \sum_{i=0}^{n_w m^2}(D(R(h, \theta)) - \hat{E})_i^2,
\end{equation}
where $h$ is the height field $\theta$ non-differentiable scene parameters and $\hat{E}$ the given irradiance of the target caustic image.

The update network directly starts with the first block of the UNet part of the network, albeit with a fixed number of channels and a different small output network that potentially contracts the channels over multiple steps and thus a larger receptive field.
It is motivated by the success of learned gradient descent methods~\cite{andrychowicz2016, flynn2019} in solving ill-posed parameter estimation problems.
The main idea is to replace the classical gradient-based local update rule
\begin{equation}
    x_{i+1} = \mathbf{S}(x_i - \alpha_i \nabla x_i)
\end{equation}
with $\alpha_i > 0$ being the step size for step $i$ and $\mathbf{S}$ being projection operators and further heuristics arising in constrained optimization~\cite{kassubeck2021} with a fully learned update rule
\begin{equation}
    x_{i+1} = x_i - U(x_i, \nabla x_i).
\end{equation}
This has the advantage that the update network can learn an appropriate prior distribution, thus avoiding unwanted local minima and taking larger adaptive steps.
In our specific case the updater is conditioned on the initial height field as well as its computed gradient and the simulated as well as the target caustic image, scaled to the same spatial resolution ($m=n$).
Independent processing of the channels as in the denoiser is not applied here, since these channels are mutually dependent of each other.
As a mapping this leads to the definition of the updater as
\begin{equation}
    U : \mathbb{R}^{n \times n} \times \mathbb{R}^{n \times n} \times \mathbb{R}^{n_w \times m \times m} \times \mathbb{R}^{n_w \times m \times m} \rightarrow \mathbb{R}^{n \times n}.
\end{equation}

Overall we consider a family of networks for both the denoising and the update parts, which are parameterized via the hyperparameters (provided in our supplementary material). 
At training time we search for the best network architecture over the parameter space defined therein.
 
\section{Experiments}\label{sec:eval}
To illustrate the demonstrate the effectiveness of N-SfC, we focus on the application of quality control in glass 3D printing. However, we must note that no sufficient number of real samples could be produced to facilitate training of denoiser and update component. Thus, we provide two carefully created synthetic datasets. 

We then use such datasets to compare our reconstructions to several state-of-the-art approaches. Finally we perform an ablation study for individual framework stages. 

%%%%%%%%%%%%%%%%%%%%%%%%%%%%%%%%%%%%%%%%%%%%%%%%%%%%%%%
\subsection{Datasets}\label{ssec:data}

We provide two synthetic datasets in addition to the \textit{test dataset}: the \textit{denoising dataset} and the \textit{updater dataset}. 
They both are rendered by drawing samples from distributions carefully chosen to represent the real data distribution. 
In particular, our height fields are fully randomly created to match the deposited filaments and are designed to accurately simulate the effect of printing a fiber on top of pre-existing ones.
The physical values for the non-differentiable scene parameters were selected to match usual process values with current technology.
The parameters regarding sample count and resolution were chosen as a compromise between quality and memory consumption. 
All details for the generation of the datasets as well as explored ranges of parameters are provided in our supplementary material. 

\begin{figure}[bh]
    \centering
    \includegraphics[width=.99\linewidth]{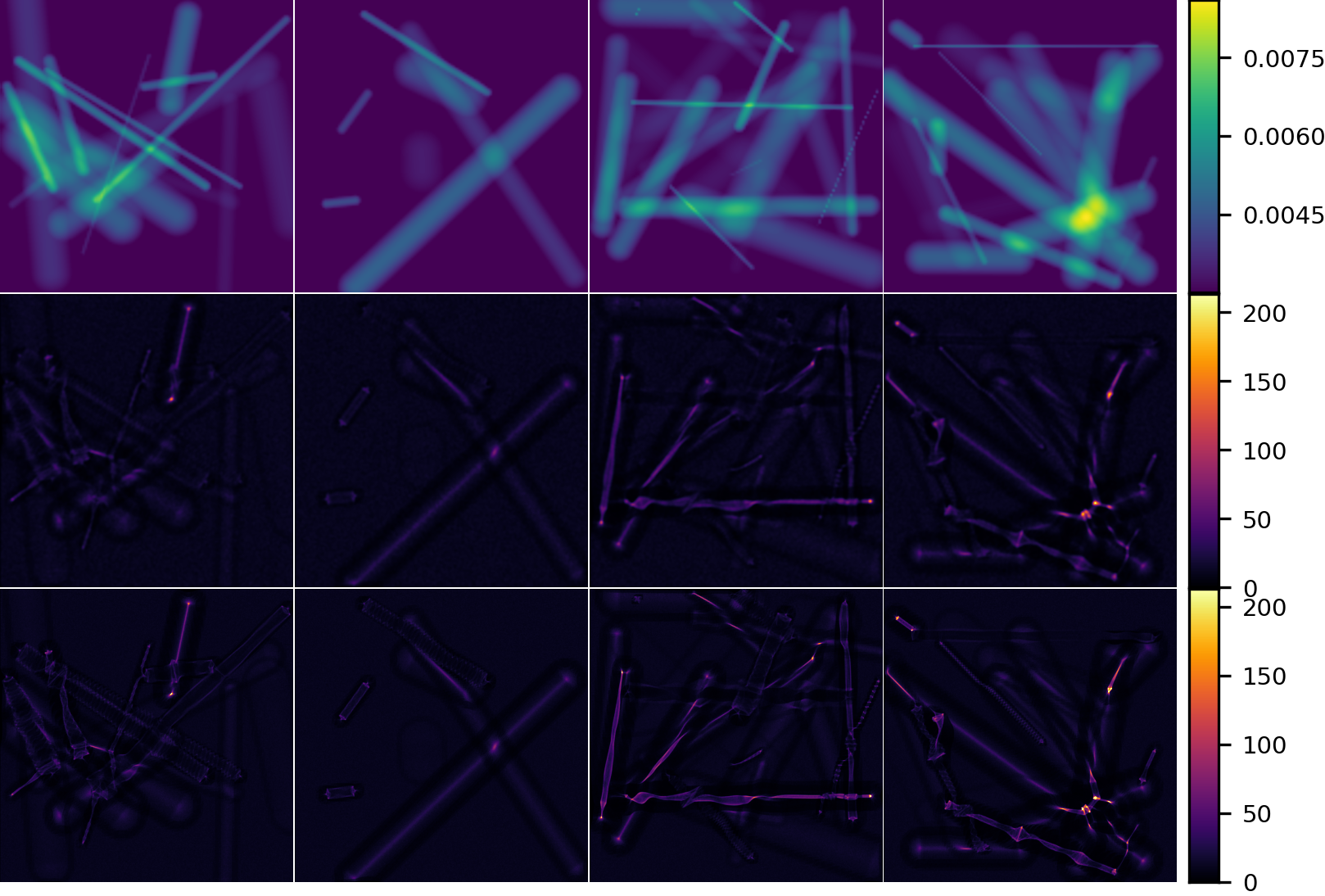}
\caption{\textbf{Denoising dataset}: we generate training data for our denoiser by sampling random ground-truth height fields from line distribution (top), which are then used to render low quality ($n_l$=$10^6$ samples, middle) and high quality ($n_l$=$1.6 \cdot 10^7$ samples, bottom) caustics images. During training, we use the low quality renderings as network inputs and the high quality versions for supervision.
    }
    \label{fig:denoising_dataset}
\end{figure}

For the \textit{denoising dataset} we draw $50000$ samples from the height field distribution and render two caustic images with different quality levels: one with $n_l$ = $10^6$ light samples and one with $n_l$ = $1.6 \cdot 10^7$ light samples. The first four entries of the dataset are shown in \cref{fig:denoising_dataset}. %
It is clearly visible that the lines create complex caustic patterns when crossing
%
%%% FIGURE DENOISING DATASET
%%
%%
%%% FIGURE UPDATER DATASET
%%
\begin{figure}[th]
    \centering
    \includegraphics[width=\linewidth]{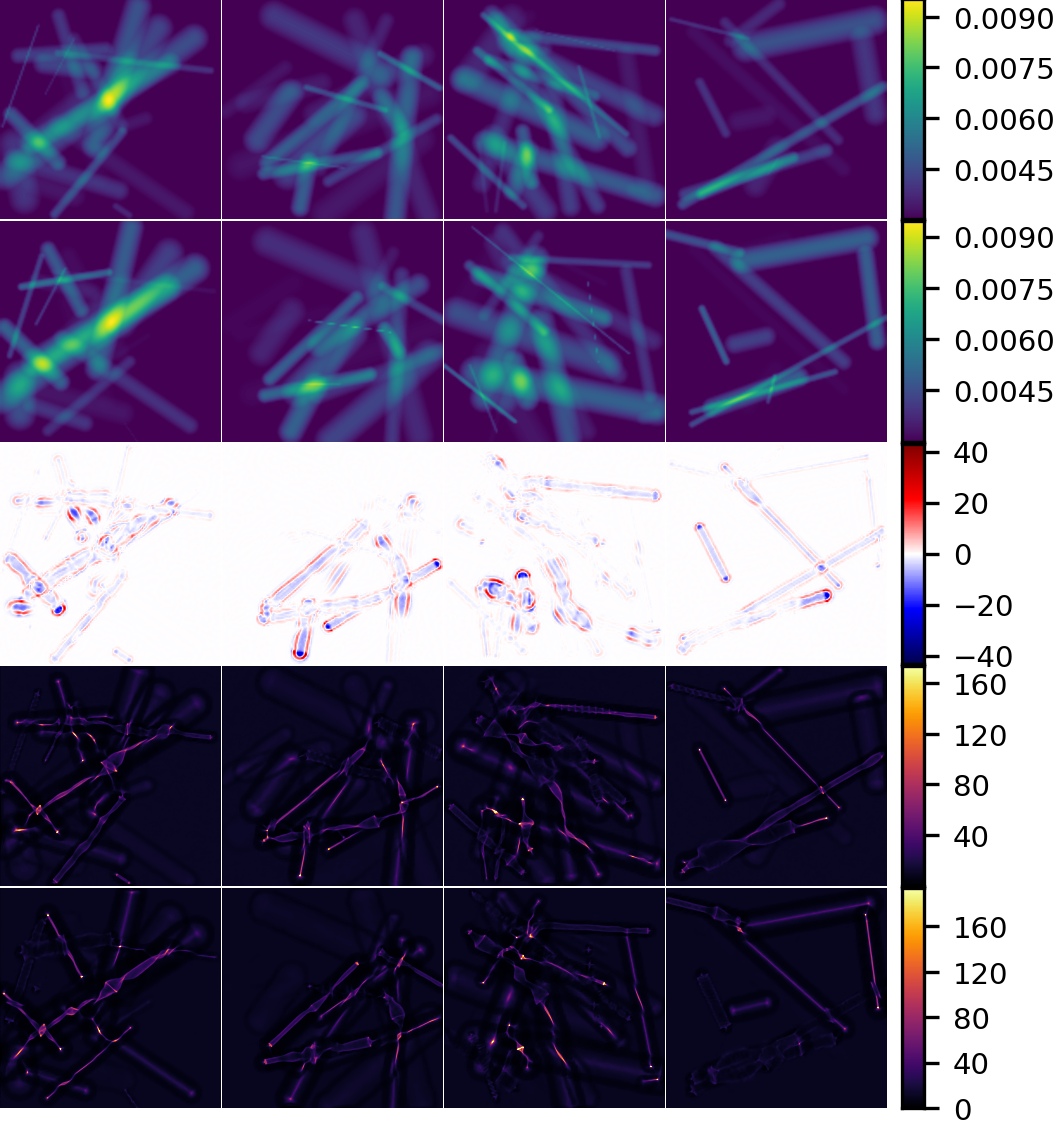}
\caption{\textbf{Updater dataset}: %similar to our denoising dataset, 
we sample pairs of source and target height fields to train our updater.
    Each sample contains (from top to bottom): the source height field, the target ground-truth height field, the height field gradient obtained via backpropagation of $\mathbf{L_{irrad}}$, as well as the simulated caustic images output by our denoiser for the source and target height fields respectively. 
    }
    \label{fig:lgd_dataset}
\end{figure}
over each other, giving the denoising network many image patches to learn representative and varied caustic patterns.

For the \textit{updater dataset} we sample two height fields from two distributions, one to represent the initial height field in the updater network,  while the other, an offset of the current estimate, is used as the target height field of the updater.
We then render caustic images for the current estimated height field and the target height field and use the trained denoiser to produce respective caustic images.
We further compute the MSE loss between the two to generate the gradient with respect to the current estimated height field.
The current height field and its caustic, the target height field and its caustic and the height field gradient are then saved for this dataset.
We generated $100000$ samples using this procedure,
the first four of which can be seen in \cref{fig:lgd_dataset}.

%%FIGURE TEST SET
%%
\begin{figure}[th]
    \centering
    \includegraphics[width=\linewidth]{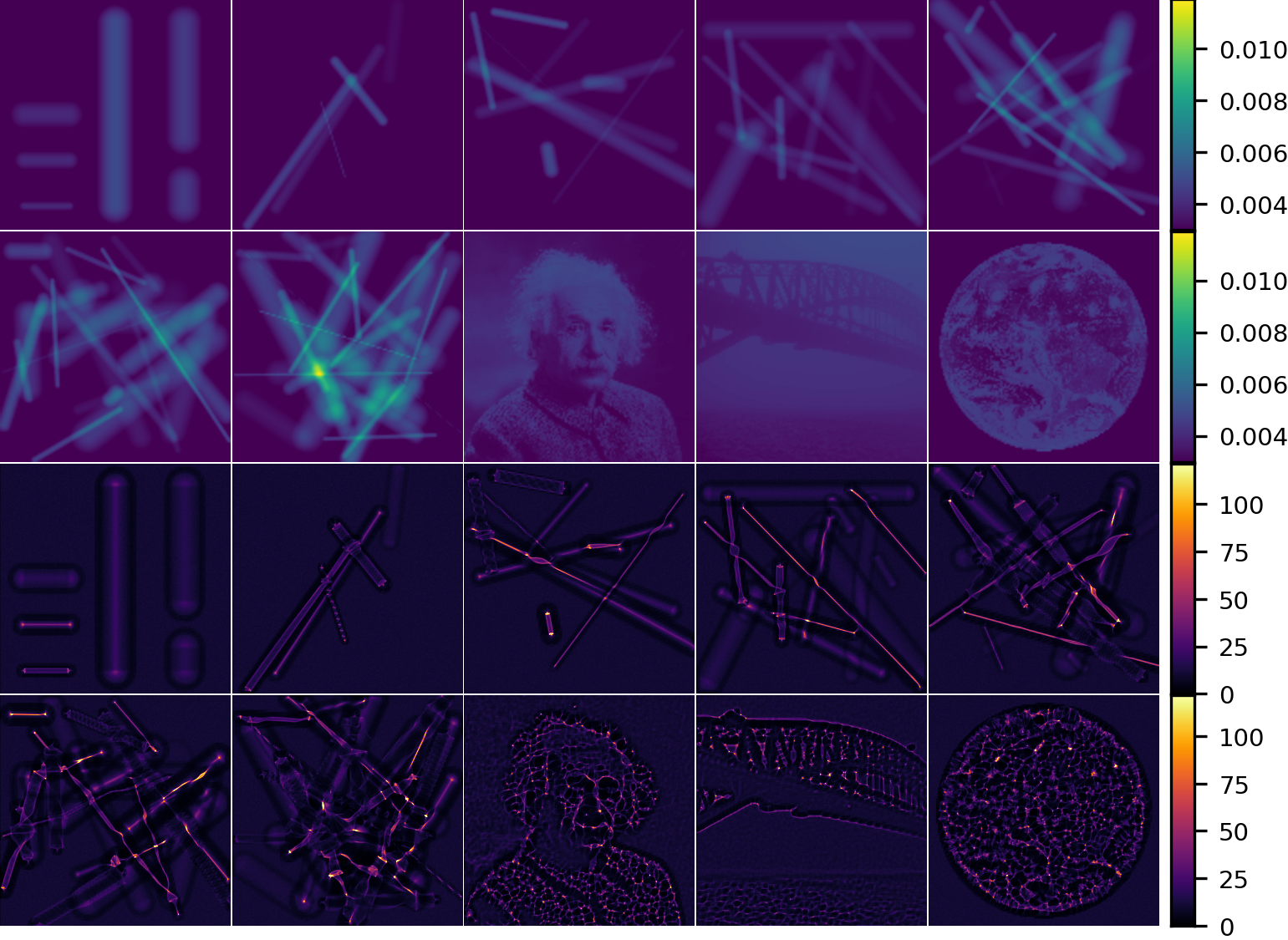}
\caption{\textbf{Test set}: we evaluate our framework on a dedicated test set containing 10 height fields of varying complexity (top). We include one sample without overlapping lines, six examples sampled randomly from the training distribution, and three grayscale photographs converted to height fields. The bottom rows show the corresponding high quality caustic images.}
    \label{fig:testset}
\end{figure}

Lastly we provide a \textit{test dataset} with 10 samples: one with hand-picked lines in the same distribution as the training data, but not present in any training data set.
Six further samples with varying complexity, \ie with 5 to 30 random lines from the same distribution as the training data, but not present in any training data set.
And finally 3 creative commons gray-scale images converted into height fields and scaled to the same value range as the training data as out-of distribution samples.
The samples and resulting high-quality ($1.6 \cdot 10^7$ light paths) caustic images are shown in \cref{fig:testset}.
\subsection{Methods and Metrics}\label{ssec:metrics}

We compare our method against SfC~\cite{kassubeck2021} and the work from Schwartzburg~\etal~\cite{schwartzburg2014}.
When comparing the accuracy of different reconstructions, we report the SSIM metric~\cite{wang2004image} as well as the \textit{relative} height field $L_2$ error
\begin{equation}
    \mathbf{L_{rel}}(h, \hat{h}) = \frac{\lVert h - \hat{h} \rVert_2}{\lVert \hat{h} \rVert_2},
\end{equation}
where $h$ is the reconstruction result and $\hat{h}$ is the ground-truth height field.
All reconstructions are initialised with a constant height field of $d=3\text{mm}$.

%%%%%%%%%%%%%%%%%%% TABLE ERRORS %%%%%%%%%%%%%%%%%%%%
\begin{table*}[ht]
    \centering
    \setlength\tabcolsep{12 pt}%11
    {\scriptsize%\footnotesize%\small
    \begin{tabular}{p{.5cm}|||p{.7cm}||c|c||c|c|||p{.7cm}||c|c||c|c|||}
        \multirow{3}{\linewidth}{\centering Test Img.} &  \multicolumn{5}{c|||}{SSIM $\uparrow$}& \multicolumn{5}{c|||}{$\mathbf{L_{rel}} \downarrow$}\\
        \cline{2-11}
          & \multirow{2}{\linewidth}{\centering \cite{schwartzburg2014}}& \multicolumn{2}{c||}{SfC} & \multicolumn{2}{c|||}{N-SfC} &\multirow{2}{\linewidth}{\centering \cite{schwartzburg2014}}& \multicolumn{2}{c||}{SfC} & \multicolumn{2}{c|||}{N-SfC}\\ \cline{3-6} \cline{8-11}
           &  & iter. 1 & iter. 3 & iter. 1 & iter. 3 & & iter. 1 & iter. 3 & iter. 1 & iter. 3\\ \hline
         \centering 1  & 0.7126 & 0.6366 & 0.1315 & 0.7297 & \textbf{0.7408} & 0.1549 & 0.1918 & 0.3566 & \textbf{0.0688} & 0.1700\\
         \centering 2  & 0.6805 & 0.6666 & 0.1448 & 0.6954 & \textbf{0.6964} & 0.0890 & 0.0991 & 0.3659 & \textbf{0.0655} & 0.0708 \\
         \centering 3  & 0.6828 & 0.6367 & 0.1228 & 0.6898 & \textbf{0.6982} & 0.1244 & 0.1333 & 0.3900 & 0.1010 & \textbf{0.0765}\\
         \centering 4  & 0.7200 & 0.6487 & 0.1136 & 0.7195 & \textbf{0.7258} & 0.1280 & 0.1379 & 0.3774 & 0.1193 & \textbf{0.0951}\\
         \centering 5  & 0.6249 & 0.5551 & 0.1105 & 0.6473 & \textbf{0.6754} & 0.2588 & 0.2613 & 0.3576 & 0.2146 & \textbf{0.1429}\\
         \centering 6  & 0.6213 & 0.5189 & 0.1093 & 0.6301 & \textbf{0.6726} & 0.2835 & 0.2887 & 0.3500 & 0.2411 & \textbf{0.1588}\\
         \centering 7  & 0.5418 & 0.4587 & 0.0986 & 0.5704 & \textbf{0.6178} & 0.3845 & 0.3832 & 0.3736 & 0.3365 & \textbf{0.2476}\\
         \centering 8  & 0.6676 & 0.6274 & 0.0845 & 0.6562 & \textbf{0.6735} & 0.2269 & 0.2133 & 0.3331 & 0.2061 & \textbf{0.1547}\\
         \centering 9  & \textbf{0.7155} & 0.6851 & 0.1010 & 0.6980 & 0.7071 & 0.2791 & 0.2700 & 0.2947 & 0.2708 & \textbf{0.2328}\\
         \centering 10 & \textbf{0.5954} & 0.5452 & 0.0909 & 0.5768 & 0.5795 & 0.2219 & 0.2081 & 0.3361 & 0.1968 & \textbf{0.1192}\\ \hline
         \centering \textit{Avg.} & \textit{0.6562}  & \textit{0.5979} & \textit{0.1108} & \textit{0.6613} & \textit{\textbf{0.6787}} & \textit{0.2151} & \textit{0.2187} & \textit{0.3535} & \textit{0.1821} & \textbf{\textit{0.1468}}
    \end{tabular}
    }
\caption{\textbf{Comparisons on the \textit{relative} height field reconstruction errors}: we compare the error of our method against SfC\cite{kassubeck2021} and Schwartzburg~\etal\cite{schwartzburg2014}, according to SSIM~\cite{wang2004image} and $\mathbf{L_{rel}}$.}
    \label{tab:errors}
\end{table*}

% SSIM
% 1,  & x.xxxx & 0.6366 & 0.1315 & 0.7297 & 0.7408
% 2,  & x.xxxx  & 0.6666 & 0.1448 & 0.6954 & 0.6964
% 3,  & x.xxxx  & 0.6367 & 0.1228 & 0.6898 & 0.6982
% 4,  & x.xxxx  & 0.6487 & 0.1136 & 0.7195 & 0.7258
% 5,  & x.xxxx  & 0.5551 & 0.1105 & 0.6473 & 0.6754
% 6,  & x.xxxx  & 0.5189 & 0.1093 & 0.6301 & 0.6726
% 7,  & x.xxxx  & 0.4587 & 0.0986 & 0.5704 & 0.6178
% 8,  & x.xxxx  & 0.6274 & 0.0845 & 0.6562 & 0.6735
% 9,  & x.xxxx  & 0.6851 & 0.1010 & 0.6980 & 0.7071
% 10,  & x.xxxx  & 0.5452 & 0.0909 & 0.5768 & 0.5795
% AVG  & x.xxxx  & 0.5979 & 0.11075 & 0.66132 & 0.67871

%%%%%%%%%%%%%%%%%%%%%%%%%%%%%%%%%%%%%%%%%%%%%%%%%%%%%%%%%%%%%
%
%
\begin{figure*}[ht]
    \centering
    \includegraphics[width=\linewidth, keepaspectratio]{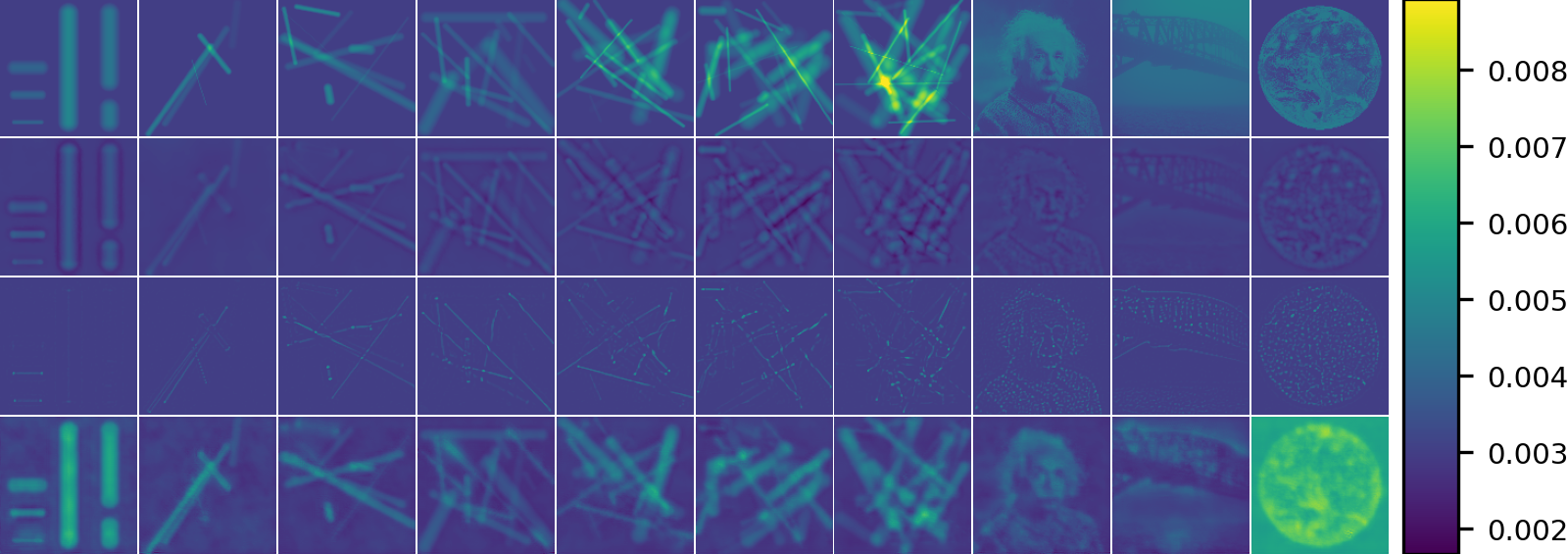}
    \caption{\textbf{Reconstruction results}: From top to bottom, ground-truth, fully converged Schwartzburg~\etal\cite{schwartzburg2014}, SfC after one iteration and N-SfC after three iterations.
        Direct face validity shows that the reference methods underestimate the actual height.
    }
    \label{fig:combined_results}
\end{figure*}
As far as parameters are concerned, N-SfC is parameter-free at test-time, whereas SfC depends on several hyperparameters, namely $\alpha_p$, $\tau_p$ and $\gamma$ for the variant with extended thresholded nonlinear Landweber scheme and volume heuristic (called M2V1 in~\cite{kassubeck2021}).
Analogous to the network parameter search we optimize these parameters on the first test sample and leave them constant for the remaining samples.
We do the same for the reconstruction parameters for \cite{schwartzburg2014}.
%%%
%
\subsection{Results and Comparisons}\label{ssec:results_comparisons}
The reconstructions on our test set for the three compared methods are depicted in \cref{fig:combined_results}, alongside with their numerical comparisons in~\cref{tab:errors} . 

From those results is clear that SfC fails to converge to a good reconstruction and is outperformed by N-SfC.
Furthermore the shape error consistently rises after one SfC update step and continues to do so until some manually defined upper height limit is reached.
We hypothesize that this is due to reduced smoothing of and subsequently sharper and higher-dynamic range caustic images of our setup, when compared to the original SfC paper.
In contrast N-SfC can decrease the shape error by repeated execution of update steps, excepting the most simple samples one and two, where the error is already low after one update step.
Despite not being trained in a recurrent manner, the updater seems to have learned a good representation of gradient descent dynamics, since for the more complex samples the error continues to decrease even after the iterations displayed in \cref{tab:errors}.
We chose to report the error after 3 steps there, since on average this produces the best result on our test data.
However, this could be further improved with an adaptive heuristic with respect to the number of iterations, as shown in the \textit{min.} column in \cref{tab:ablation}.
With respect to Schwartzburg~\etal's method we can see that it generally outperforms SfC in most cases, the exception being SfC after 1 iteration in $\mathbf{L_{rel}}$ metric.
However, on average and in most samples it is still outperformed by our full N-SfC method.

\begin{table*}[ht]
    \centering
    \setlength\tabcolsep{10 pt}%9
    {\scriptsize%\footnotesize%\scriptsize%
    \begin{tabular}{p{.5cm}|||c|c||c|c|||c||c|c||c|c|c||c|||}
        \multirow{3}{\linewidth}{\centering Test Img.} &  \multicolumn{4}{c|||}{SSIM $\uparrow$}& \multicolumn{7}{c|||}{$\mathbf{L_{rel}} \downarrow$}\\
        \cline{2-12}
         & \multicolumn{2}{c||}{N-SfC w/o den.} & \multicolumn{2}{c|||}{N-SfC w/o grad.} & Initial & \multicolumn{2}{c||}{N-SfC w/o den.} & \multicolumn{3}{c||}{N-SfC w/o grad.} & N-SfC\\ \cline{2-5} \cline{7-12}
         & iter. 1 & iter. 3 & iter. 1 & iter. 3 & error & iter. 1 & iter. 3 & iter. 1 & iter. 3 & min. & min.\\ \hline
         \centering 1  & 0.6483 & 0.5393 & 0.7371 & 0.7275 & 0.1954 & 0.1041 & 0.2222 & 0.0643 & 0.1582 & 0.0643 & 0.0688\\
         \centering 2  & 0.5702 & 0.4420 & 0.7005 & 0.6966 & 0.1102 & 0.1268 & 0.2549 & 0.0532 & 0.0667 & 0.0532 & 0.0614\\
         \centering 3  & 0.5791 & 0.4558 & 0.6968 & 0.6985 & 0.1477 & 0.1362 & 1.0318 & 0.0888 & 0.0774 & 0.0774 & \textbf{0.0765}\\
         \centering 4 & 0.5769 & 0.4476 & 0.7302 & 0.7289 & 0.1538 & 0.1500 & 0.2727 & 0.1054 & 0.0951 & 0.0946 & \textbf{0.0898}\\
         \centering 5  & 0.5112 & 0.3942 & 0.6613 & 0.6797 & 0.2770 & 0.2470 & 0.2877 & 0.1971 & 0.1398 & 0.1151 & \textbf{0.1104}\\
         \centering 6  & 0.5073 & 0.4044 & 0.6490 & 0.6791 & 0.3061 & 0.2721 & 0.2706 & 0.2227 & 0.1488 & 0.1163 & 0.1195\\
         \centering 7 & 0.4501 & 0.3653 & 0.5856 & 0.6210 & 0.3984 & 0.3689 & 0.3862 & 0.3252 & 0.2457 & 0.1629 & 0.1651\\
         \centering 8 & 0.5254 & 0.3824 & 0.6673 & 0.6774 & 0.2346 & 0.2413 & 0.2591 & 0.1953 & 0.1512 & 0.1246 & \textbf{0.1162}\\
         \centering 9 & 0.5610 & 0.4059 & 0.7043 & 0.7062 & 0.2823 & 0.2875 & 0.3010 & 0.2629 & 0.2325 & 0.1646 & \textbf{0.1451}\\
         \centering 10 & 0.4631 & 0.3522 & 0.5980 & 0.6119 & 0.2372 & 0.2336 & 0.2376 & 0.1782 & 0.1154 & 0.0971 & \textbf{0.0938}\\ \hline
         \centering\textit{Avg.} & \textit{0.5393} & \textit{0.4189} & \textit{0.6730} & \textit{0.6827} & \textit{0.2343} &\textit{ 0.2168} & \textit{0.3524} & \textit{0.1693} & \textit{0.1431} & \textit{0.1070}&\textbf{0.1047}

    \end{tabular}
    }
    \caption{\textbf{Ablation} of \textit{relative} height field reconstruction errors, according to SSIM~\cite{wang2004image} and $\mathbf{L_{rel}}$.}
    \label{tab:ablation}
\end{table*}

% & 0.6483 & 0.5393 & 0.7371 & 0.7275
% & 0.5702 & 0.4420 & 0.7005 & 0.6966
% & 0.5791 & 0.4558 & 0.6968 & 0.6985
% & 0.5769 & 0.4476 & 0.7302 & 0.7289
% & 0.5112 & 0.3942 & 0.6613 & 0.6797
% & 0.5073 & 0.4044 & 0.6490 & 0.6791
% & 0.4501 & 0.3653 & 0.5856 & 0.6210
% & 0.5254 & 0.3824 & 0.6673 & 0.6774
% & 0.5610 & 0.4059 & 0.7043 & 0.7062
% & 0.4631 & 0.3522 & 0.5980 & 0.6119
%& \textit{0.5393} & \textit{0.4189} & \textit{0.6730} & \textit{0.6827}

% 0,0.6482842,0.5392802,0.7370679,0.7274716

% 1,0.5702127,0.44196677,0.700517,0.69662297

% 2,0.57911575,0.45579022,0.69675565,0.698496

% 3,0.5769276,0.44760686,0.73020077,0.72892535

% 4,0.51124567,0.39421564,0.661348,0.67973125

% 5,0.50727403,0.40439424,0.6490027,0.6790813

% 6,0.45010325,0.365305,0.5856075,0.6209557

% 7,0.5254183,0.38240033,0.66733897,0.677371

% 8,0.5609652,0.4059388,0.70428556,0.7062051

% 9,0.46314028,0.35220388,0.598004,0.6119109

%%%%%%%%%%%%%%%%%%%%%%%%%%%%%%%%%%%%%%%%%%%%%%%%%%%%%%%%%%%%%
Lastly we wish to address runtimes of our method compared to SfC and Schwartzburg~\etal's method.
This is not a trivial comparison, since the former failed to converge in our test cases and the latter is only present as a CPU implementation.
However the authors of SfC reported a runtime of 2.95 minutes for 72 iterations on a height field, which is structurally quite similar to our first test set sample. 
All samples, when reconstructed with Schwartzburg~\etal's method took more than 15 minutes with several hundred iterations each, which would not benefit much from GPU acceleration due to its sequential nature.
We however created all of our reconstructions in only one to three iterations with an average iteration time of 1.59 seconds per iteration, which is an improvement factor of 37 to 111 in speed compared to SfC.
%
%%%%%%%%%%%%%%%%%%%%%%%%%%%%%%%%%%%%%%%%%%%%%%%%%%%%%%%%%%%%%%%%%%%%%%%%%%%%%%
\subsection{Ablation Study}\label{ssec:ablation_study} 

In the following, we ablate parts of our processing to illustrate the influence of single components.
We trained a variant of our best performing updater network with the direct output from our render module, leaving out the denoiser.
We denote this variant as \textit{N-SfC w/o den.} in \cref{tab:ablation} 
and the numerical results reveal that this variation significantly under performs our full method.
In some cases this variation even increases the initial shape error, leading to worse performance than SfC.
This demonstrates the importance of the denoising component in the reconstruction context.

Another ablation we performed is training an updater where the gradient of the height field as computed by our differentiable rendering framework is replaced with zeros. 
We denote this variant as \textit{N-SfC w/o grad.} in \cref{tab:ablation}, 
which has more subtle influence than \textit{N-SfC w/o den}.
When comparing the reconstruction results after one and 3 iterations against those from our full method (see \cref{tab:errors}), this variant outperforms the full network, where the gradient is set to some meaningful input.
However comparing the minimal surface error reached over all iterations in columns \textit{min.} in \cref{tab:ablation} reveals that the gradient has a meaningful and positive influence on the minimal reachable reconstruction error, especially for the more complex out-of-distribution test samples (8 - 10). 
The samples, with minimal all-iteration error below the ablated all-iteration error are marked in bold.
From this observation we can conclude two things: 
On the one hand, our method can be trained and executed without the presence of a differentiable renderer, only the forward caustic image simulation is strictly necessary.
On the other hand to get the best possible reconstruction results, gradients from a differentiable renderer help in most cases, though they only contribute very local information (see \cref{ssec:underdeterminism}).
However they have to be combined with a good stopping criterion for the updater, which performs the right amount of update steps for the height field complexity to be reconstructed.

\section{Discussion and Limitations}\label{sec:dis} 
The above experiments show that our method outperforms the current state-of-the art by a significant margin. 
At the same time, our current implementation and datasets make several assumptions on the concrete area of application, thus implying some implicit limitations. 
As previously mentioned, our fully trained model is completely free of any additional parameters, which avoids the need for manual hyperparameter tuning, including the choice of step sizes. 
However the applied training data and learning process are adjusted towards the physical scene configuration (such as the light and screen positions) of our \textit{in-situ} manufacturing setup. 
Generating more diverse training data and providing the networks with explicit knowledge about certain scene parameters could help to further improve the generalizability of our model.

In the ablation study, we found that passing the gradient from a differentiable rendering module does not necessarily improve the reconstruction error within the first gradient descent steps, which is in line with recent research (see \cite{morris2005} and \cref{fig:underdeterminism}), stating that local information can be counterproductive for settings like ours.
However, further analysis revealed that local gradient information can still come in handy to converge towards a better minimum (and thus achieve a lower error) during later iterations.
In order to fully benefit from these findings, our framework could  further be extended with a mechanism for choosing an appropriate stopping criterion, that cannot be implemented using only the caustic observations in our current setup due to their sensitivity to noise in the underlying height fields.

%%%%%%%%%%%%%%%%%%%%%%%%%%%%%%%%%%%%%%%%%%%%%%%%%%%%%%%%%%%%%%%%%%%%%%%%%%%%%%%%%%%%%%%%%%%%%%%%%%%%%%%%%%%%%%%%%%%%%%%%%%
\section{Conclusion}\label{sec:conclusion} 
We presented Neural-Shape from Caustics (N-SfC), a fast and flexible method for reconstructing the shape of translucent objects from a single caustic image.
We combine recent work on differentiable light transport simulation with our novel neural denoising components and learned gradient descent optimizer to significantly improve both the stability and quality of the iterative reconstruction process.
Our quantitative and qualitative analysis showed that our neural approach outperforms current state-of-the-art approaches in terms of final reconstruction error and compute requirements. 
Furthermore, we found that our learned gradient-based update scheme enables better generalization and overall flexibility, making the approach adaptable to practical applications such as integrated quality feedback control for glass manufacturing processes.
\iftoggle{cvprfinal}{\\}{\vfill}
 
\iftoggle{cvprfinal}{\noindent\textbf{Our supplementary material} contains additional details on the network architectures, the datasets as well as comparisons and visualizations of other techniques. 
The full implementation is available at our project page {\small\url{https://graphics.tu-bs.de/publications/kassubeck2021n-sfc}}.

\noindent\textbf{Acknowledgements:} 
This work was partially funded by the DFG under Germany’s Excellence Strategy within the Cluster of Excellence PhoenixD (EXC 2122, Project ID 390833453).
\iftoggle{cvprfinal}{\vfill}{}}{}
%%\input{TemplatesLatex/guidelines}
%%%%%%%%%%%%%%%%%%%%%%%%%%%%%%%%%%%%%%%%%%%%%%%%%%%%%%%%%%%%%%%%%%%%%%%%%%%%%%
%%%%%%%%% REFERENCES
{\small
\bibliographystyle{ieee_fullname}
\bibliography{references}

\begin{thebibliography}{10}\itemsep=-1pt

\bibitem{andrychowicz2016}
Marcin Andrychowicz, Misha Denil, Sergio Gomez, Matthew~W Hoffman, David Pfau,
  Tom Schaul, Brendan Shillingford, and Nando De~Freitas.
\newblock Learning to learn by gradient descent by gradient descent.
\newblock In {\em Advances in neural information processing systems}, pages
  3981--3989, 2016.

\bibitem{arnold2021}
Autodesk.
\newblock Denoiser, 2021.
\newblock https://docs.arnoldrenderer.com/
  display/A5AF3DSUG/Denoiser\#Denoiser-OptiXDenoiser [Online; Last accessed
  01-11-2021.

\bibitem{bako2017}
Steve Bako, Thijs Vogels, Brian McWilliams, Mark Meyer, Jan Nov{\'a}k, Alex
  Harvill, Pradeep Sen, Tony Derose, and Fabrice Rousselle.
\newblock Kernel-predicting convolutional networks for denoising monte carlo
  renderings.
\newblock {\em ACM Trans. Graph.}, 36(4):97--1, 2017.

\bibitem{bangaru2020}
Sai Bangaru, Tzu-Mao Li, and Fr{\'e}do Durand.
\newblock Unbiased warped-area sampling for differentiable rendering.
\newblock {\em ACM Trans. Graph.}, 39(6):245:1--245:18, 2020.

\bibitem{wandb}
Lukas Biewald.
\newblock Experiment tracking with weights and biases, 2020.
\newblock Software available from wandb.com.

\bibitem{chaitanya2017}
Chakravarty R~Alla Chaitanya, Anton~S Kaplanyan, Christoph Schied, Marco Salvi,
  Aaron Lefohn, Derek Nowrouzezahrai, and Timo Aila.
\newblock Interactive reconstruction of monte carlo image sequences using a
  recurrent denoising autoencoder.
\newblock {\em ACM Transactions on Graphics (TOG)}, 36(4):1--12, 2017.

\bibitem{pytorch-lightning}
William {Falcon et al.}
\newblock Pytorch lightning, 2019.
\newblock Software available from pytorchlightning.ai.

\bibitem{flynn2019}
John Flynn, Michael Broxton, Paul Debevec, Matthew DuVall, Graham Fyffe, Ryan
  Overbeck, Noah Snavely, and Richard Tucker.
\newblock Deepview: View synthesis with learned gradient descent.
\newblock In {\em Proceedings of the IEEE/CVF Conference on Computer Vision and
  Pattern Recognition}, pages 2367--2376, 2019.

\bibitem{frisvad2014}
J.R. Frisvad, L. Schj{\o}th, K. Erleben, and J. Sporring.
\newblock Photon differential splatting for rendering caustics.
\newblock {\em Computer Graphics Forum}, 33(6):252--263, Sept. 2014.

\bibitem{Hochreiter2001}
Sepp Hochreiter, A.~Steven Younger, and Peter~R. Conwell.
\newblock Learning to learn using gradient descent.
\newblock In Georg Dorffner, Horst Bischof, and Kurt Hornik, editors, {\em
  Artificial Neural Networks --- ICANN 2001}, pages 87--94, Berlin, Heidelberg,
  2001. Springer Berlin Heidelberg.

\bibitem{huo2021}
Yuchi Huo and Sung-eui Yoon.
\newblock A survey on deep learning-based monte carlo denoising.
\newblock {\em Computational Visual Media}, 7(2):169--185, 2021.

\bibitem{jiang2021}
Giulio Jiang and Bernhard Kainz.
\newblock Deep radiance caching: Convolutional autoencoders deeper in ray
  tracing.
\newblock {\em Computers \& Graphics}, 94:22--31, 2021.

\bibitem{kalantari2015}
Nima~Khademi Kalantari, Steve Bako, and Pradeep Sen.
\newblock A machine learning approach for filtering monte carlo noise.
\newblock {\em ACM Trans. Graph.}, 34(4):122--1, 2015.

\bibitem{kassubeck2021}
Marc Kassubeck, Florian B{\"u}rgel, Susana Castillo, Sebastian Stiller, and
  Marcus Magnor.
\newblock Shape from caustics: Reconstruction of 3d-printed glass from
  simulated caustic images.
\newblock In {\em IEEE WACV}, pages 2877--2886, Jan 2021.

\bibitem{kingma2017adam}
Diederik~P. Kingma and Jimmy Ba.
\newblock Adam: A method for stochastic optimization, 2017.

\bibitem{li2017}
Lisha Li, Kevin Jamieson, Giulia DeSalvo, Afshin Rostamizadeh, and Ameet
  Talwalkar.
\newblock Hyperband: A novel bandit-based approach to hyperparameter
  optimization.
\newblock {\em The Journal of Machine Learning Research}, 18(1):6765--6816,
  2017.

\bibitem{li2018}
Tzu-Mao Li, Miika Aittala, Fr{\'e}do Durand, and Jaakko Lehtinen.
\newblock Differentiable monte carlo ray tracing through edge sampling.
\newblock {\em ACM Trans. Graph. (Proc. SIGGRAPH Asia)}, 37(6):222:1--222:11,
  2018.

\bibitem{loubet2019}
G. Loubet, N. Holzschuch, and W. Jakob.
\newblock Reparameterizing discontinuous integrands for differentiable
  rendering.
\newblock {\em Transactions on Graphics (Proceedings of SIGGRAPH Asia)},
  38(6):2019, 2019.

\bibitem{luan2021}
Fujun Luan, Shuang Zhao, Kavita Bala, and Zhao Dong.
\newblock Unified shape and svbrdf recovery using differentiable monte carlo
  rendering.
\newblock {\em Computer Graphics Forum}, 40(4):101--113, 2021.

\bibitem{lyu2020}
Jiahui Lyu, Bojian Wu, Dani Lischinski, Daniel Cohen-Or, and Hui Huang.
\newblock Differentiable refraction-tracing for mesh reconstruction of
  transparent objects.
\newblock {\em ACM Transactions on Graphics (Proceedings of SIGGRAPH ASIA
  2020)}, 39(6):195:1--195:13, 2020.

\bibitem{meyron2018}
Jocelyn Meyron, Quentin M{\'e}rigot, and Boris Thibert.
\newblock Light in power: a general and parameter-free algorithm for caustic
  design.
\newblock {\em ACM Trans. Graph.}, 37(6):1--13, Dec. 2018.

\bibitem{morris2005}
N.J.W. Morris and K.N. Kutulakos.
\newblock Dynamic refraction stereo.
\newblock In {\em Tenth IEEE International Conference on Computer Vision
  (ICCV'05) Volume 1}, volume~2, pages 1573--1580 Vol. 2, 2005.

\bibitem{SuperCaustics}
Mehdi Mousavi and Rolando Estrada.
\newblock Supercaustics: Real-time, open-source simulation of transparent
  objects for deep learning applications.
\newblock {\em CoRR}, abs/2107.11008, 2021.

\bibitem{nalbach2017}
O. Nalbach, E. Arabadzhiyska, D. Mehta, H.-P. Seidel, and T. Ritschel.
\newblock Deep shading: Convolutional neural networks for screen space shading.
\newblock {\em Comput. Graph. Forum}, 36(4):65–78, jul 2017.

\bibitem{nimier2020}
Merlin Nimier-David, S{\'e}bastien Speierer, Beno{\^\i}t Ruiz, and Wenzel
  Jakob.
\newblock Radiative backpropagation: an adjoint method for lightning-fast
  differentiable rendering.
\newblock {\em ACM Transactions on Graphics (TOG)}, 39(4):146--1, 2020.

\bibitem{nimier2019}
M. Nimier-David, D. Vicini, T. Zeltner, and W. Jakob.
\newblock Mitsuba 2: A retargetable forward and inverse renderer.
\newblock {\em Transactions on Graphics (Proceedings of SIGGRAPH Asia)}, 38(6),
  2019.

\bibitem{paszke2017}
A. Paszke, S. Gross, S. Chintala, G. Chanan, E. Yang, Z. DeVito, Z. Lin, A.
  Desmaison, L. Antiga, and A. Lerer.
\newblock Automatic differentiation in pytorch.
\newblock In {\em NIPS 2017 Workshop on Autodiff}, 2017.

\bibitem{paszke2019}
A. Paszke, S. Gross, F. Massa, A. Lerer, J. Bradbury, G. Chanan, T. Killeen, Z.
  Lin, N. Gimelshein, L. Antiga, A. Desmaison, A. Kopf, E. Yang, Z. DeVito, M.
  Raison, A. Tejani, S. Chilamkurthy, B. Steiner, L. Fang, J. Bai, and S.
  Chintala.
\newblock Pytorch: An imperative style, high-performance deep learning library.
\newblock In H. Wallach, H. Larochelle, A. Beygelzimer, F. d\textquotesingle
  Alch\'{e}-Buc, E. Fox, and R. Garnett, editors, {\em Advances in Neural
  Information Processing Systems 32}, pages 8024--8035. Curran Associates,
  Inc., 2019.

\bibitem{UNet2015}
Olaf Ronneberger, Philipp Fischer, and Thomas Brox.
\newblock U-net: Convolutional networks for biomedical image segmentation.
\newblock In Nassir Navab, Joachim Hornegger, William~M. Wells, and
  Alejandro~F. Frangi, editors, {\em Medical Image Computing and
  Computer-Assisted Intervention -- MICCAI 2015}, pages 234--241, Cham, 2015.
  Springer International Publishing.

\bibitem{Schmidhuber1992}
Jürgen Schmidhuber.
\newblock Learning to control fast-weight memories: An alternative to dynamic
  recurrent networks.
\newblock {\em Neural Computation}, 4(1):131--139, 1992.

\bibitem{Schmidhuber1993}
J. Schmidhuber.
\newblock A neural network that embeds its own meta-levels.
\newblock In {\em IEEE International Conference on Neural Networks}, pages
  407--412 vol.1, 1993.

\bibitem{Schmidhuber1997}
Jürgen Schmidhuber, Jieyu Zhao, and Marco Wiering.
\newblock Shifting inductive bias with success-story algorithm, adaptive levin
  search, and incremental self-improvement.
\newblock {\em Machine Learning}, 28:105--130, 01 1997.

\bibitem{schwartzburg2014}
Y. Schwartzburg, R. Testuz, A. Tagliasacchi, and {others}.
\newblock High-contrast computational caustic design.
\newblock {\em ACM Transactions on Graphics}, 2014.

\bibitem{snoek2012}
Jasper Snoek, Hugo Larochelle, and Ryan~P Adams.
\newblock Practical bayesian optimization of machine learning algorithms.
\newblock {\em Advances in neural information processing systems}, 25, 2012.

\bibitem{vicini2021}
Delio Vicini, S{\'e}bastien Speierer, and Wenzel Jakob.
\newblock Path replay backpropagation: differentiating light paths using
  constant memory and linear time.
\newblock {\em ACM Transactions on Graphics (TOG)}, 40(4):1--14, 2021.

\bibitem{vogels2018}
Thijs Vogels, Fabrice Rousselle, Brian McWilliams, Gerhard R{\"o}thlin, Alex
  Harvill, David Adler, Mark Meyer, and Jan Nov{\'a}k.
\newblock Denoising with kernel prediction and asymmetric loss functions.
\newblock {\em ACM Transactions on Graphics (TOG)}, 37(4):1--15, 2018.

\bibitem{Von-Witzendorff2018}
P. von Witzendorff, L. Pohl, O. Suttmann, P. Heinrich, A. Heinrich, J. Zander,
  H. Bragard, and S. Kaierle.
\newblock Additive manufacturing of glass: {CO2-Laser} glass deposition
  printing.
\newblock {\em Procedia CIRP}, 74:272--275, Jan. 2018.

\bibitem{wang2004image}
Zhou Wang, Alan~C. Bovik, Hamid~R. Sheikh, and Eero~P. Simoncelli.
\newblock Image quality assessment: from error visibility to structural
  similarity.
\newblock {\em IEEE Transactions on Image Processing}, 13(4):600--612, 2004.

\bibitem{Wetzstein2011}
Gordon Wetzstein, David Roodnick, Wolfgang Heidrich, and Ramesh Raskar.
\newblock Refractive shape from light field distortion.
\newblock In {\em 2011 International Conference on Computer Vision}, pages
  1180--1186, 2011.

\bibitem{Younger2001}
A.S. Younger, S. Hochreiter, and P.R. Conwell.
\newblock Meta-learning with backpropagation.
\newblock In {\em IJCNN'01. International Joint Conference on Neural Networks.
  Proceedings (Cat. No.01CH37222)}, volume~3, pages 2001--2006 vol.3, 2001.

\bibitem{zeng2020}
Zheng Zeng, Lu Wang, Bei-Bei Wang, Chun-Meng Kang, and Yan-Ning Xu.
\newblock Denoising stochastic progressive photon mapping renderings using a
  multi-residual network.
\newblock {\em Journal of Computer Science and Technology}, 35:506--521, 2020.

\bibitem{zhang2020}
Cheng Zhang, Bailey Miller, Kai Yan, Ioannis Gkioulekas, and Shuang Zhao.
\newblock Path-space differentiable rendering.
\newblock {\em ACM Trans. Graph.}, 39(4):143:1--143:19, 2020.

\bibitem{zhang2021}
Cheng Zhang, Zihan Yu, and Shuang Zhao.
\newblock Path-space differentiable rendering of participating media.
\newblock {\em ACM Trans. Graph.}, 40(4):76:1--76:15, 2021.

\bibitem{CVPRTut2021}
Shuang Zhao, Ioannis Gkioulekas, and Sai Bangaru.
\newblock {CVPR} tutorial on physics-based differentiable rendering, 2021.
\newblock https://www.diff-render.org/.

\bibitem{zhu2020}
Shilin Zhu, Zexiang Xu, Henrik~Wann Jensen, Hao Su, and Ravi Ramamoorthi.
\newblock Deep kernel density estimation for photon mapping.
\newblock {\em Computer Graphics Forum}, 39(4):35--45, 2020.

\end{thebibliography}
}
%%%%%%%%%%%%%%%%%%%%%%%%%%%%%%%%%%%%%%%%%%%%%%%%%%%%%%%%%%%%%%%%%%%%%%%%%%%%%%
\twocolumn[{ 
\renewcommand\twocolumn[1][]{#1} 
\subtitle{N-SfC: Robust and Fast Shape Estimation from Caustic Images \\ --- Supplementary Material ---}
}]
\noindent In this supplementary document, we provide more details on the proposed methodology given in the main paper, implementation details along with additional results for the presented framework. \\
We will also provide the full code including dataset generation scripts upon publication of this paper. We acknowledge the positive change CVPR is attempting with regard to providing review versions of code and dataset artifacts. However have to refrain from doing so, because creating a version of the code with all author references removed is significant effort in itself and uploading it without any copyright whatsoever puts a significant portion of this work at risk in our opinion.
\section{Network Architecture} 
\begin{figure*}
    \centering
    \includegraphics[width=\textwidth]{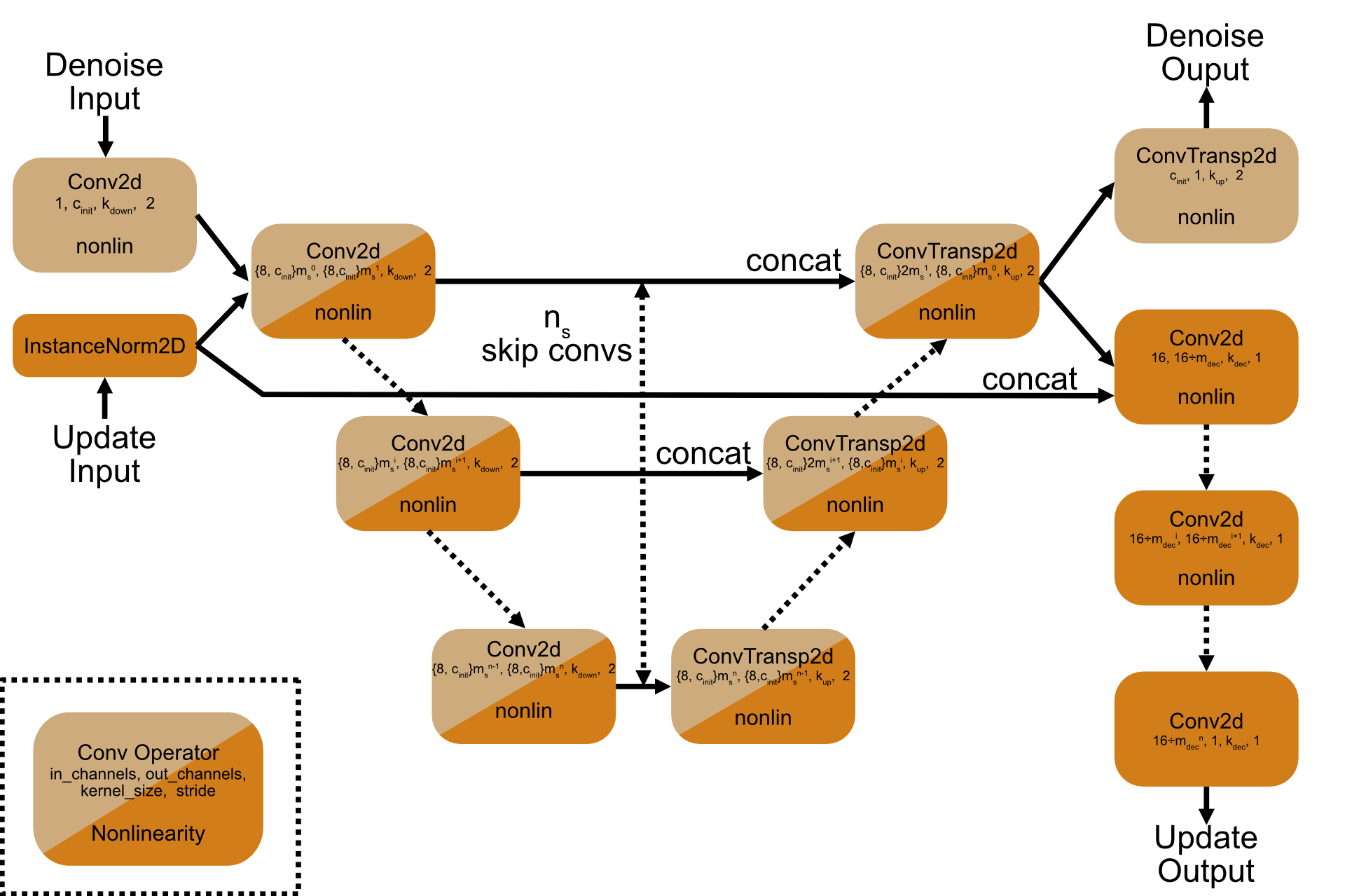}
    \caption{\textbf{The network architectures} are variants of the UNet~\cite{UNet2015} architecture. 
    The denoiser and updater network differ mainly in their input and output computations. 
    The coloring depicts, whether a block belongs to denoiser or updater respectively. }
    \label{fig:networks}
\end{figure*}

Subsequent processing of caustic images is dependent on two neural-network components, which belong to the same architectural family as depicted in~\cref{fig:networks}.
Both networks share a structure similar to UNet~\cite{UNet2015} as the main component and differ in a few blocks with respect to the input and output. 
The denoiser includes a single \texttt{Conv + nonlin} block, which expands the number of channels to $c_{init}\in [1, 32]$ channels for the UNet part of the network.
An equivalent block contracts those channels after the UNet part of the denoising network.
Note that the number of input channels in the denoising network is given as 1, even though the caustic images are produced with $n_w$ channels, meaning each channel is denoised independently, such that this network could potentially handle simulation input with as few as one or many spectral channels.

Overall we consider a family of networks for both the denoising and the update parts, which are parameterized via the hyperparameters listed in \cref{tab:network_parameters}.
At training time we search for the best network architecture over the parameter space defined therein.

\begin{table*}[ht]
    \centering
    {\footnotesize
    \begin{tabular}{l|l|l|l|c|c}
         Parameter & Meaning & Type & Search Value & Best Denoiser & Best Updater\\ \hline
         {\scriptsize\texttt{learning rate}} & Learning rate of optimizer & float & $\in [0.0001, 0.1]$ & 0.00151 & 0.005155 \\
         $c_{init}$ & Initial number of channels for UNet & integer & $\in [1, 32]$ & 31 & - \\
         $nonlin$ & Specific nonlinearity to employ after convolutions & -- &{\scriptsize {\texttt{ELU}, \texttt{ReLU}, \texttt{PReLU}, \texttt{SELU}}} & {\scriptsize\texttt{PReLU}} & {\scriptsize\texttt{PReLU}} \\
         $k_{down}$ & Kernel size in downconvolution part of network & integer & $\in [2, 11]$ & 5 & 9 \\
         $k_{up}$ & Kernel size in upconvolution part of network & integer & $\in [2, 11]$ & 2 & 9 \\
         $m_s$ & Channel multiplier for each depth layer of UNet & integer & $\in [1, 8]$ & 2 & 8 \\
         $n_s$ & Number of skip connections; depth of UNet & integer & $\in [1, 4]$ & 4 & 1 \\
         $m_{dec}$ & Channel divisor in each output block & integer & $\in [2, 11]$ & - & 8 \\
         $k_{dec}$ & Kernel size in each output block & integer & $\in \{2, 4, 8, 16\}$ & - & 4 
    \end{tabular}
    }
    \caption{Parameters for family of networks.}
    \label{tab:network_parameters}
\end{table*}

\section{Dataset and Training Details}
Given our focus on the application of quality control in glass 3D printing, all physical values represent usual values with current technology.

As mentioned in the main paper, it is crucial that the training data accurately samples the distribution of real values. Thus, to define parameters for our dataset of glass substrates it it beneficial to recall some parameters of the underlying production process.

The process works by depositing glass fibers of diameter of roughly $0.4$mm onto $3$mm thick glass substrates, for which an area of $5 \times 5$ cm is common.
In our simplified setup, only the source and the screen surface interact with the light paths. All parameters regarding scene structure can be found in \cref{tab:scene_parameters}.
The parameters regarding sample count and resolution were chosen as a compromise between quality and memory consumption.
For more details about the influence of each rendering parameter we refer to~\cite{frisvad2014}.
\begin{table*}[ht]
    \centering
    {\small
    \begin{tabular}{l|l|l}
         Parameter & Meaning & Value \\ \hline
         $n \times n$ & Pixel resolution of height map & (128, 128) \\
         $m \times m$ & Pixel resolution of caustic image & (512, 512) \\
         $d \in \mathbf{R}$ & Base thickness of substrate & 3mm \\
         $n_l \in \mathbf{N}$ & Number of samples, \ie light paths & $1e^6$ and $16e^6$ \\
         $n_w \in \mathbf{N}$ & Number of wavelengths in simulation & 3: \{610nm, 530nm, 430nm\} \\
         $s \in \mathbf{R}$ & Smoothing parameter for photon footprint  & 16.0 \\
         $\alpha \in \mathbf{R}_{+}$ & Angle of light emission & 0 (collimated light) \\
         $L_i \in \mathbf{R}_{+}^{n_w}$ & Radiosity of light source & (1 W/m$^2$, 1 W/m$^2$, 1 W/m$^2$) \\
         $L_p \in \mathbf{R}^3$ & Position of light source & (0m, 0m, 1m) \\
         $S_p \in \mathbf{R}^3$ & Position of screen & (0m, 0m, $-1e^{-6}$m) \\
         $(h_x, h_y) \in \mathbf{R}_{+}^2$ & Size of the base substrate & (5cm, 5cm)
    \end{tabular}
    }
    \caption{\textbf{Non-Differentiable Scene Parameters}. 
    Note that \textbf{non-differentiable} does not indicate an intrinsic limitation of the approach, but is meant to indicate that we do not optimize for these parameters and thus chose not to compute gradients for them.
    The last column denotes the values considered for the scene setup in dataset creation and the final optimization loops. }
    \label{tab:scene_parameters}
\end{table*}

Our height fields are randomly create using an arbitrary number of straight line segments to match the deposited filaments.
A random number of lines with aleatory startpoints, endpoints, width and height are drawn. 
Those lines have a cosine falloff from their center to the edge of their width and to get a final height field we add these lines together to simulate the effect of printing a fiber on top of pre-existing ones.
The samples are uniformly drawn from the ranges defined in \cref{tab:dataset_ranges}.
For better illustration we also provide the dataset figures from the main paper in \cref{fig:denoising_dataset:supp}, \cref{fig:lgd_dataset:supp} and \cref{fig:testset:supp} when discussing the details of their creation.

\begin{table}[ht]
    \centering
    \begin{tabular}{l|l|l}
         Parameter & Value Range & Offsets\\ \hline
         Number of lines & [2, 30] & None \\
         Line Start & [-2.5cm, 2.5cm]$^2$ & [-2.5mm, 2.5mm]$^2$ \\
         Line End & [-2.5cm, 2.5cm]$^2$ & [-2.5mm, 2.5mm]$^2$ \\
         Line Width & [0.1mm, 4mm] & [-1mm, 1mm] \\
         Line Height & [0.1mm, 2mm] & [-1mm, 1mm]
    \end{tabular}
    \caption{Parameters for height fields generation}
    \label{tab:dataset_ranges}
\end{table}

\subsection{Denoising Dataset}
\begin{figure}
    \centering
    \includegraphics[width=\linewidth]{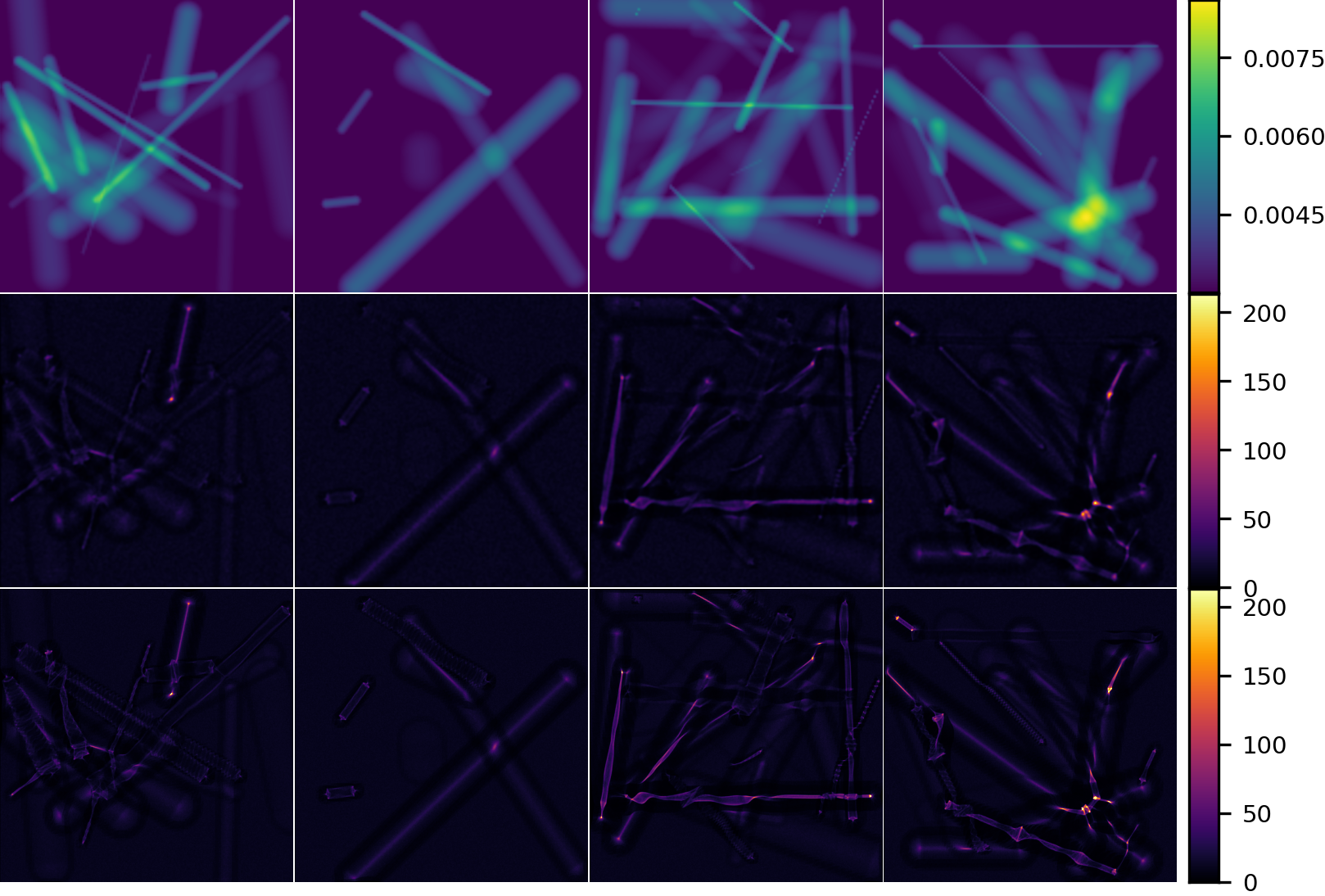}
    \caption{Denoising dataset samples}
    \label{fig:denoising_dataset:supp}
\end{figure}

For the \textit{denoising dataset} we draw $50000$ samples from this distribution and render two caustic images with different quality levels: one with $10^6$ light samples and one with $1.6 \cdot 10^7$ light samples and parameters as in \cref{tab:scene_parameters}.
The first four entries of the dataset are shown in \cref{fig:denoising_dataset:supp}.
It is clearly visible that crossing the lines create complex caustic patterns when crossing over each other, giving the denoising network many image patches to learn representative caustic patterns.

This dataset was created on a machine with a Xeon-E5 1630 CPU and two NVidia Titan RTX GPUs with 24Gb memory each, where it took roughly 17h.

\subsection{Denoiser Training}
We implemented the denoising component of the network using PyTorch Lightning~\cite{pytorch-lightning} and trained with this dataset, split randomly into 90\% training and 10\% validation images.
The loss function for this training was the MSE loss between estimated denoised caustic and the caustic image with $1.6 \cdot 10^7$ samples.
We searched over the hyperparameters as defined in \cref{tab:network_parameters} by minimizing this loss over the validation set to find the best performing architecture.
This search was performed using the hyperparameter sweeps function in~\cite{wandb} with their implementation of bayesian search~\cite{snoek2012} and hyperband early termination~\cite{li2017} with a minimum of 5 epochs. 
All models were trained with Adam optimizer~\cite{kingma2017adam} until the validation loss showed no improvement over the last 3 epochs.
The total compute time of the search was 5 days with 60 trained models over two machines. 
One being the afore mentioned Xeon-E5 1630 CPU and two NVidia Titan RTX GPUs with 24Gb memory and the other equipped with a Xeon-E5 1630 CPU and two NVidia RTX 3090 GPUs with 24Gb memory each.
The parameters of the best performing model are listed in \cref{tab:network_parameters} and were used in all following experiments.

\subsection{Updater Dataset}\label{ssec:updater_dataset}
\begin{figure}
    \centering
    \includegraphics[width=\linewidth]{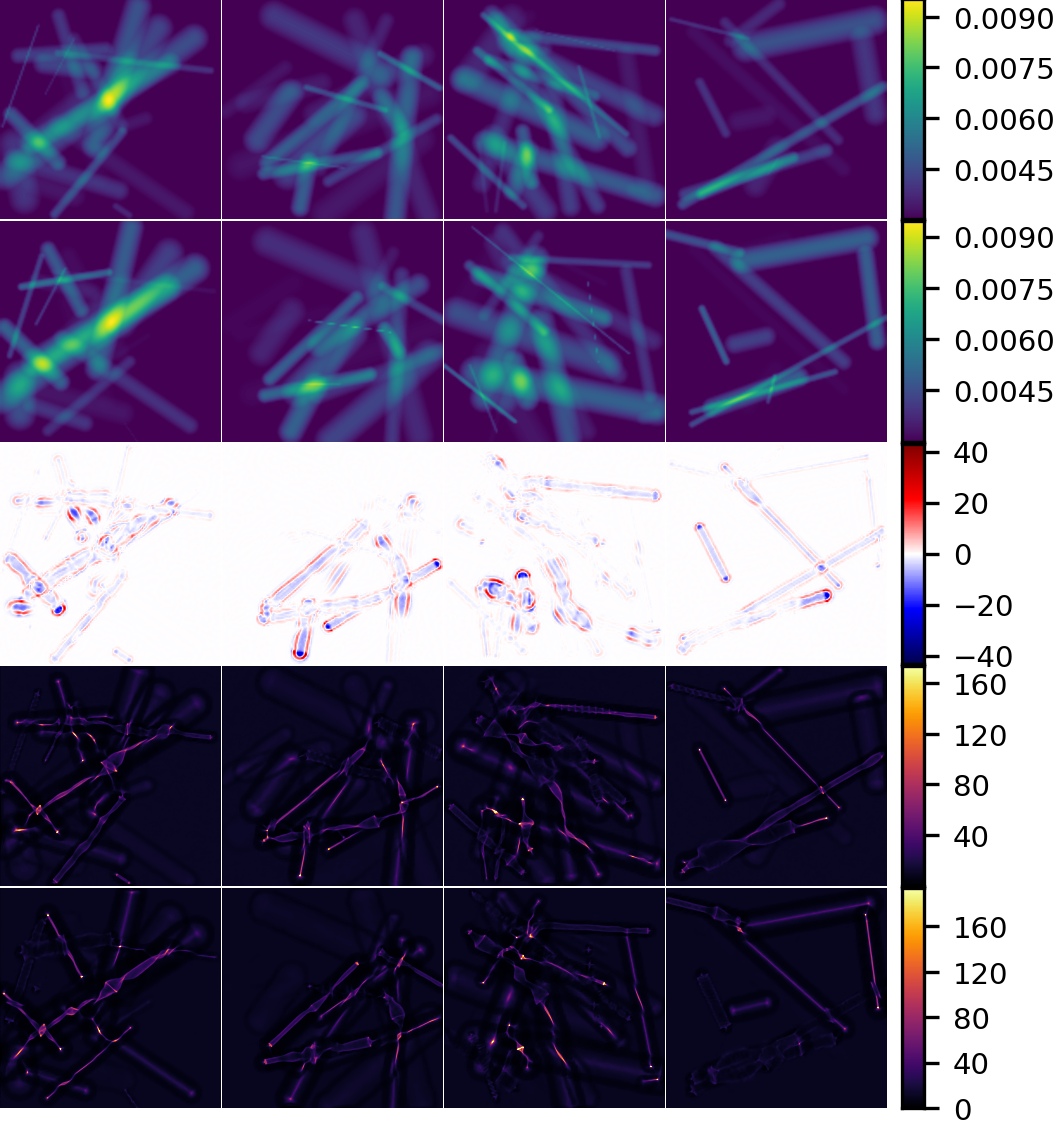}
    \caption{Updater dataset samples}
    \label{fig:lgd_dataset:supp}
\end{figure}

For the \textit{updater dataset} we sample two height fields from \cref{tab:dataset_ranges}. 
One drawn uniformly from the \textit{Value Range} column is used to represent the current estimated height field the updater network.
The other is uniformly drawn as an offset of the current estimate as depicted in the \textit{Offsets} column. 
This is then used as the target height field for the updater network.
We then render caustic images for the current estimated height field and the target height field and use the trained denoiser to produce respective caustic images.
We further compute the MSE loss between the two to generate the gradient with respect to the current estimated height field.
The current height field and its caustic, the target height field and its caustic and the height field gradient are then saved for this dataset.
We generated 100000 samples using this procedure, which took about 30h on a Xeon-E5 1630 CPU and two NVidia RTX 3090 GPUs with 24Gb memory each.
The first four of which can be seen in \cref{fig:lgd_dataset:supp}.

\subsection{Updater Training}
The reasoning behind the sampling in \cref{ssec:updater_dataset} is that learned gradient descent schemes are usually trained in an unrolled fashion with multiple consecutive update steps being supervised by ground-truth data.
Unfortunately it proved to be computationally infeasible to generate a new simulation and gradient after each update step while simultaneously optimizing for the best network architecture.
Thus we opted to create a larger dataset and train the updater point-wise by comparing the resulting height field to the target height field after a single step.
We hypothesized that this dataset contains enough varied image patches to let the network learn the dynamics of a learned gradient descent scheme and generalize to repeated execution at test time.
The rest of the training framework like dataset split, computational setup, etc., is the same as for the denoising network.
Total compute time was 20 days with 89 trained models, with best parameters as reported in \cref{tab:network_parameters}.

\subsection{Test Dataset}

\begin{figure}
    \centering
    \includegraphics[width=\linewidth]{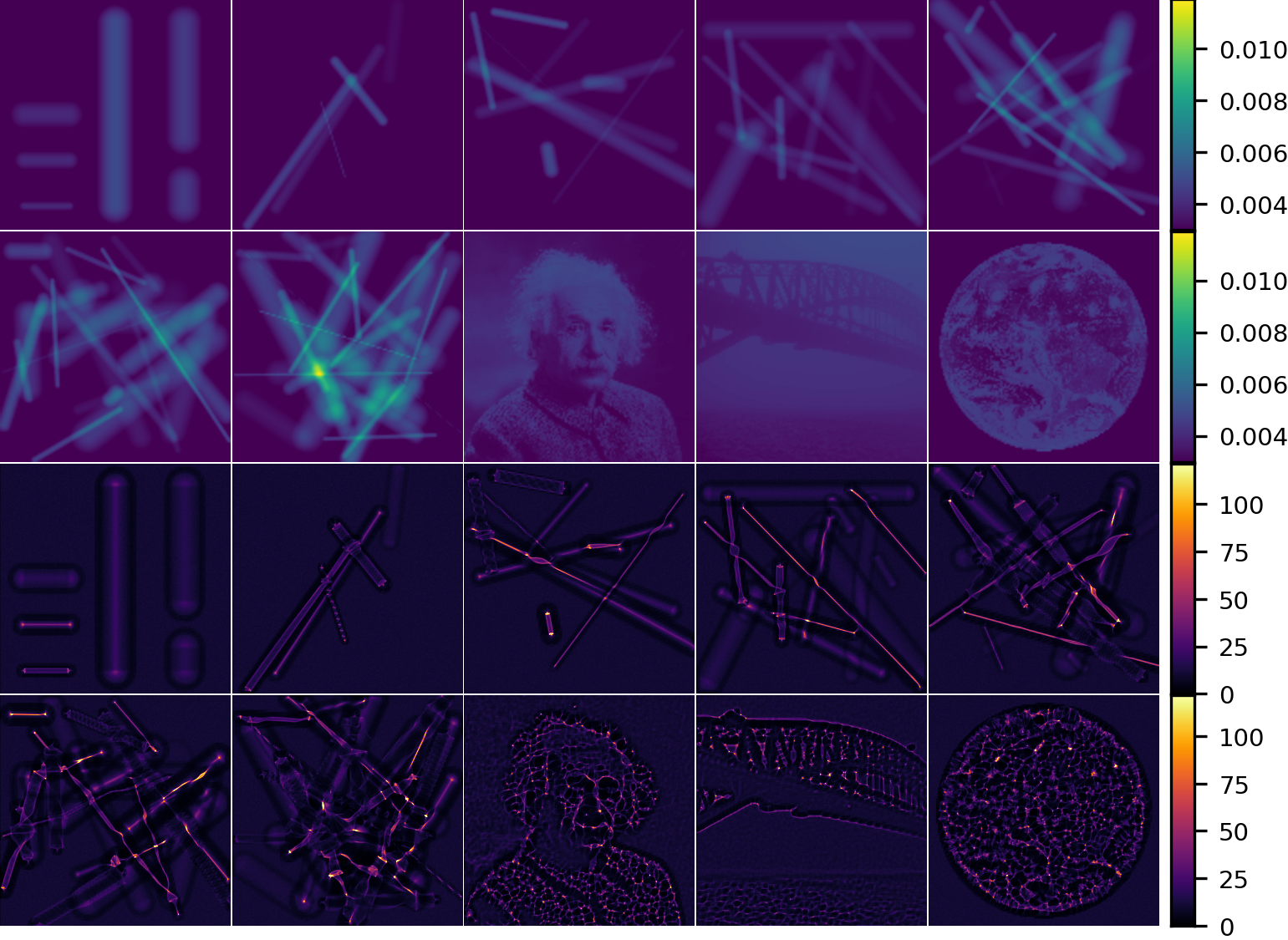}
    \caption{Test set.}
    \label{fig:testset:supp}
\end{figure}

\begin{figure*}
    \centering
    \includegraphics[width=0.17\textwidth]{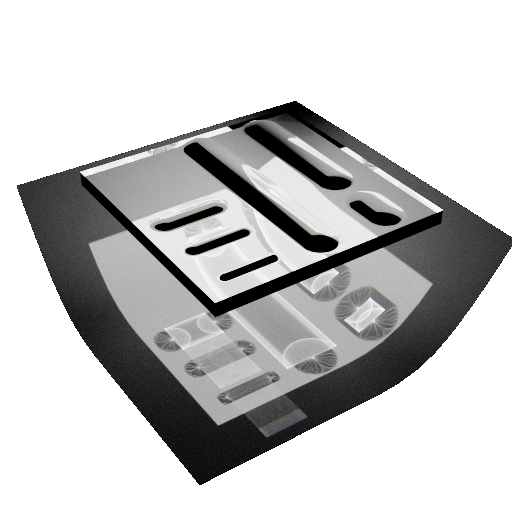}
    \includegraphics[width=0.17\textwidth]{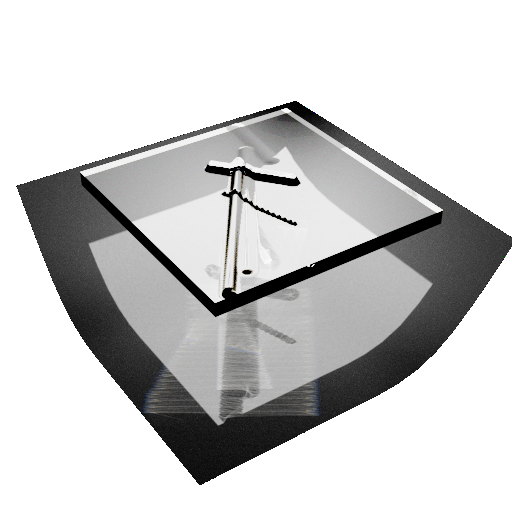}
    \includegraphics[width=0.17\textwidth]{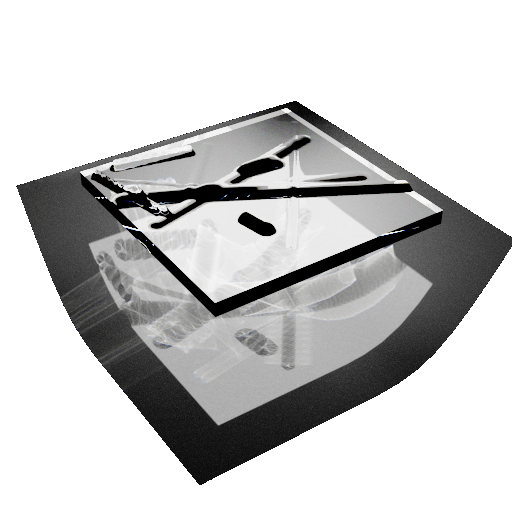}
    \includegraphics[width=0.17\textwidth]{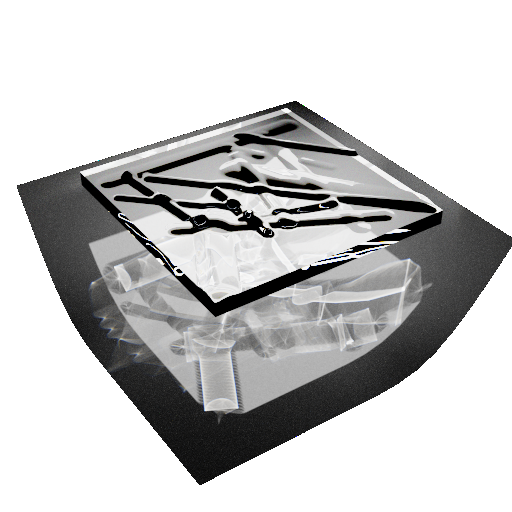}
    \includegraphics[width=0.17\textwidth]{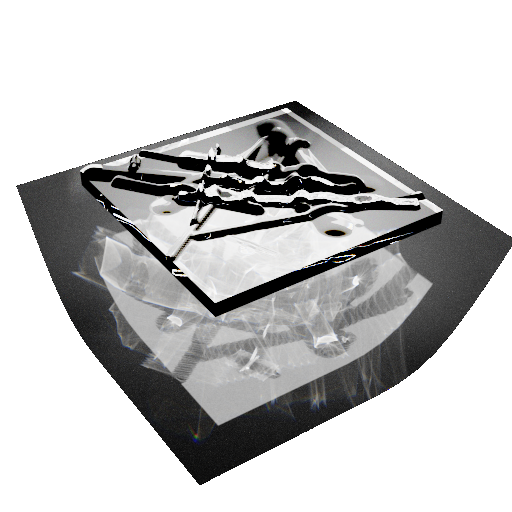}\\
    \includegraphics[width=0.17\textwidth]{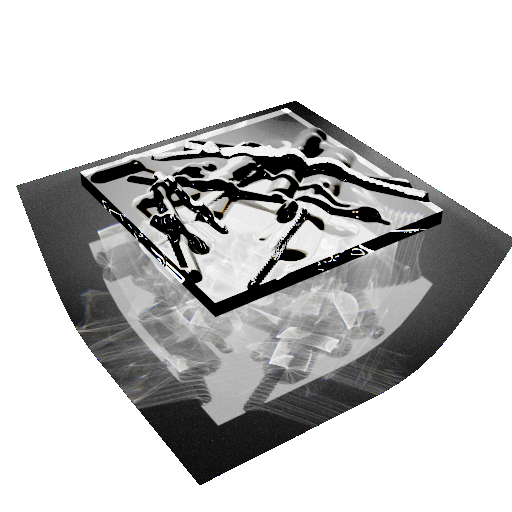}
    \includegraphics[width=0.17\textwidth]{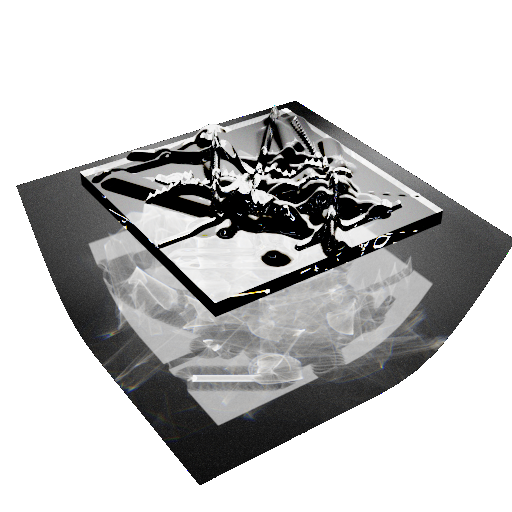}
    \includegraphics[width=0.17\textwidth]{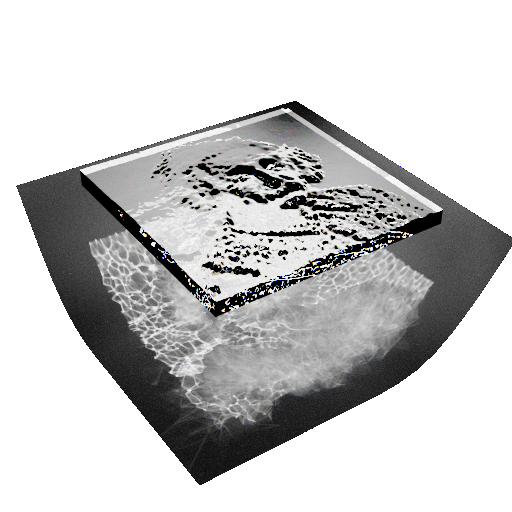}
    \includegraphics[width=0.17\textwidth]{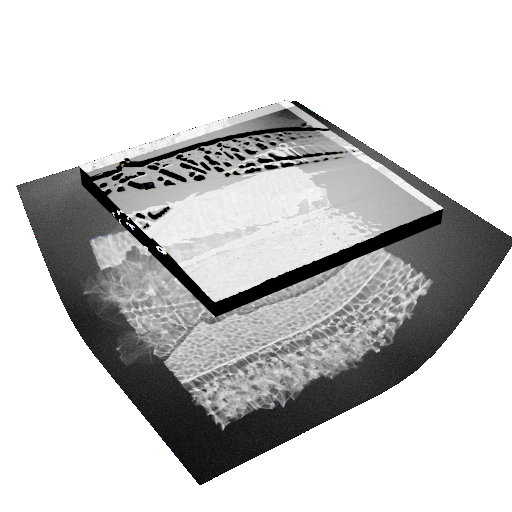}
    \includegraphics[width=0.17\textwidth]{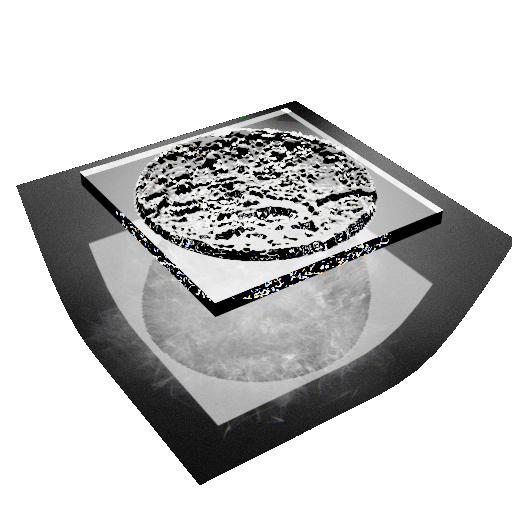}
    \caption{\textbf{Ground Truth renderings} of the test set. }
    \label{fig:testset_renderings}
\end{figure*}

Lastly we provide a \textit{test dataset} with 10 samples: one with hand-picked lines in the same distribution as the training data (see \cref{tab:dataset_ranges}), but not present in any training data set.
Six further samples with varying complexity, \ie with 5 to 30 random lines from the same distribution as the training data, but not present in any training data set.
And finally 3 creative commons gray-scale images converted into height fields and scaled to the same value range as the training data as out-of distribution samples.
The samples and resulting high-quality ($1.6 \cdot 10^7$ light paths) caustic images are shown in \cref{fig:testset:supp}.
We provide renderings of the test samples in a scene, which has been modified for better visibility in \cref{fig:testset_renderings}.

\section{Comparisons}
\begin{figure*}[th!]
    \centering
    \begin{tikzpicture}
    % viridis colormap colors
    \definecolor{clr1}{RGB}{68, 1, 84, 255}
    \definecolor{clr2}{RGB}{72, 35, 116}
    \definecolor{clr3}{RGB}{64, 67, 135}
    \definecolor{clr4}{RGB}{52, 94, 141}
    \definecolor{clr5}{RGB}{41, 120, 142}
    \definecolor{clr6}{RGB}{32, 144, 140}
    \definecolor{clr7}{RGB}{34, 167, 132}
    \definecolor{clr8}{RGB}{68, 190, 112}
    \definecolor{clr9}{RGB}{121, 209, 81}
    \definecolor{clr10}{RGB}{189, 222, 38}
    \begin{axis}[
        width=\textwidth,
        height=0.5\textwidth,
        xlabel={Iteration},
        ylabel={Relative shape error},
        xmin=0, xmax=14,
        ymin=0, ymax=0.4,
        legend columns=5,
        legend style={at={(0, 1.05)}, anchor=south west}
    ]
    
    \addplot[color=clr1, densely dashed] table[x=Step, y=rel_error, col sep=comma]{plots/sfc_0.csv};
    \addlegendentry{SfC \#1};
        
    \addplot[ color=clr2, densely dashed, ] table[x=Step, y=rel_error, col sep=comma]{plots/sfc_1.csv};
    \addlegendentry{SfC \#2};
    
    \addplot[ color=clr3, densely dashed, ] table[x=Step, y=rel_error, col sep=comma]{plots/sfc_2.csv};
    \addlegendentry{SfC \#3};
    
    \addplot[ color=clr4, densely dashed, ] table[x=Step, y=rel_error, col sep=comma]{plots/sfc_3.csv};
    \addlegendentry{SfC \#4};
    
    \addplot[ color=clr5, densely dashed, ] table[x=Step, y=rel_error, col sep=comma]{plots/sfc_4.csv};
    \addlegendentry{SfC \#5};
    
    \addplot[ color=clr6, densely dashed, ] table[x=Step, y=rel_error, col sep=comma]{plots/sfc_5.csv};
    \addlegendentry{SfC \#6};
    
    \addplot[ color=clr7, densely dashed, ] table[x=Step, y=rel_error, col sep=comma]{plots/sfc_6.csv};
    \addlegendentry{SfC \#7};
    
    \addplot[ color=clr8, densely dashed, ] table[x=Step, y=rel_error, col sep=comma]{plots/sfc_7.csv};
    \addlegendentry{SfC \#8};
    
    \addplot[ color=clr9, densely dashed, ] table[x=Step, y=rel_error, col sep=comma]{plots/sfc_8.csv};
    \addlegendentry{SfC \#9};
    
    \addplot[ color=clr10, densely dashed, ] table[x=Step, y=rel_error, col sep=comma]{plots/sfc_9.csv};
    \addlegendentry{SfC \#10};
    
    % start of nsfc
    
    \addplot[ color=clr1, ] table[x=Step, y=rel_error, col sep=comma]{plots/nsfc_0.csv};
    \addlegendentry{N-SfC \#1};
    
    \addplot[ color=clr2, ] table[x=Step, y=rel_error, col sep=comma]{plots/nsfc_1.csv};
    \addlegendentry{N-SfC \#2};
    
    \addplot[ color=clr3, ] table[x=Step, y=rel_error, col sep=comma]{plots/nsfc_2.csv};
    \addlegendentry{N-SfC \#3};
    
    \addplot[ color=clr4, ] table[x=Step, y=rel_error, col sep=comma]{plots/nsfc_3.csv};
    \addlegendentry{N-SfC \#4};
    
    \addplot[ color=clr5, ] table[x=Step, y=rel_error, col sep=comma]{plots/nsfc_4.csv};
    \addlegendentry{N-SfC \#5};
    
    \addplot[ color=clr6, ] table[x=Step, y=rel_error, col sep=comma]{plots/nsfc_5.csv};
    \addlegendentry{N-SfC \#6};
    
    \addplot[ color=clr7, ] table[x=Step, y=rel_error, col sep=comma]{plots/nsfc_6.csv};
    \addlegendentry{N-SfC \#7};
    
    \addplot[ color=clr8, ] table[x=Step, y=rel_error, col sep=comma]{plots/nsfc_7.csv};
    \addlegendentry{N-SfC \#8};
    
    \addplot[ color=clr9, ] table[x=Step, y=rel_error, col sep=comma]{plots/nsfc_8.csv};
    \addlegendentry{N-SfC \#9};
    
    \addplot[ color=clr10, ] table[x=Step, y=rel_error, col sep=comma]{plots/nsfc_9.csv};
    \addlegendentry{N-SfC \#10};
        
    % legend too big to show
    % \legend{};
    \end{axis}
    \end{tikzpicture}
    \caption{Relative shape error comparison for all test samples for SfC (dashed lines) compared to N-SfC (solid lines) during the updater iterations.
    Corresponding dataset samples are colored equally and start with the same relative error at iteration 0.
    }
    \label{fig:sfc_nsfc_convergence}
\end{figure*}
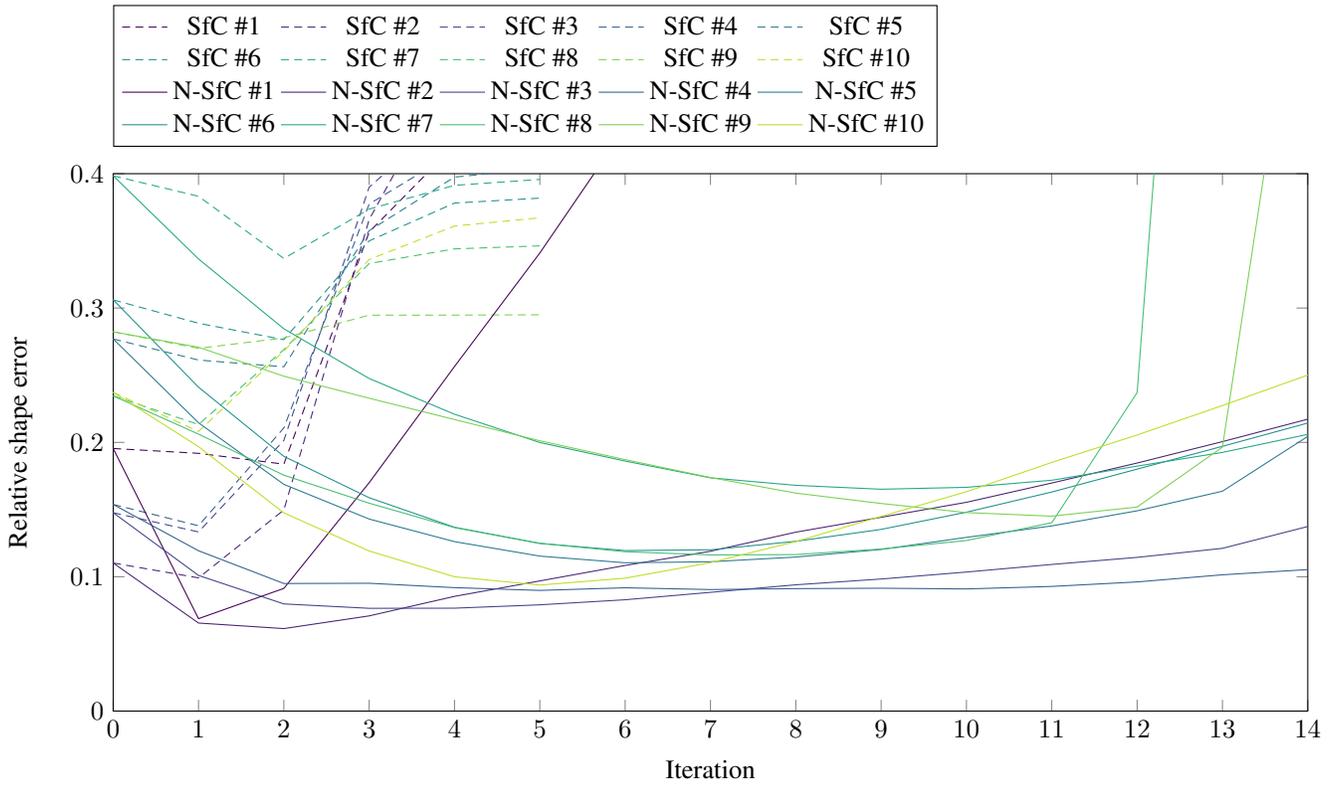

\begin{figure*}
    \centering
    \includegraphics[width=0.17\textwidth]{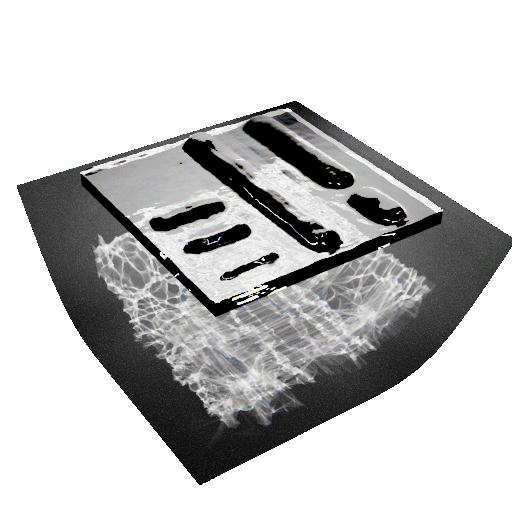}
    \includegraphics[width=0.17\textwidth]{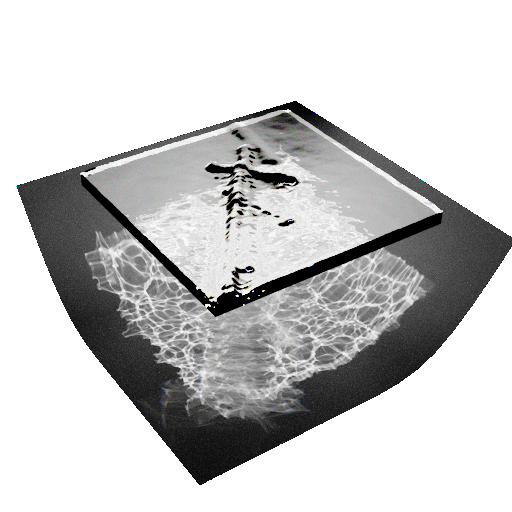}
    \includegraphics[width=0.17\textwidth]{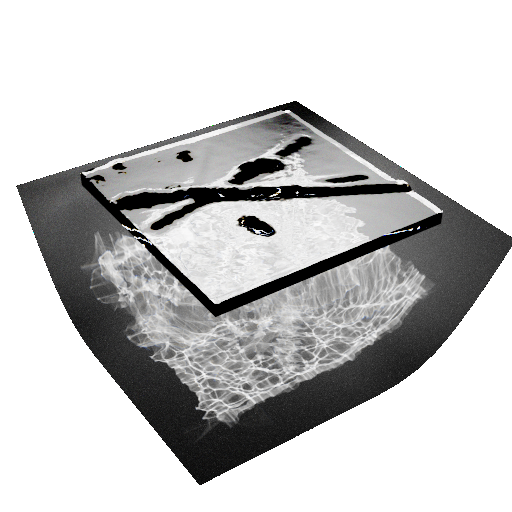}
    \includegraphics[width=0.17\textwidth]{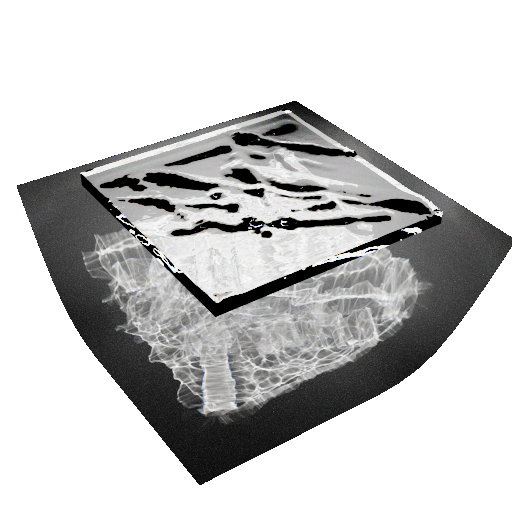}
    \includegraphics[width=0.17\textwidth]{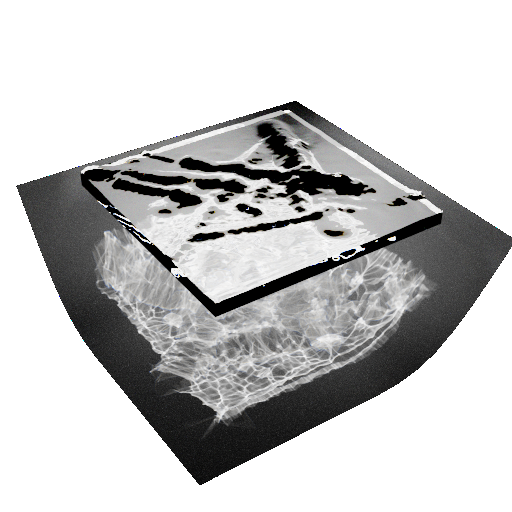}\\
    \includegraphics[width=0.17\textwidth]{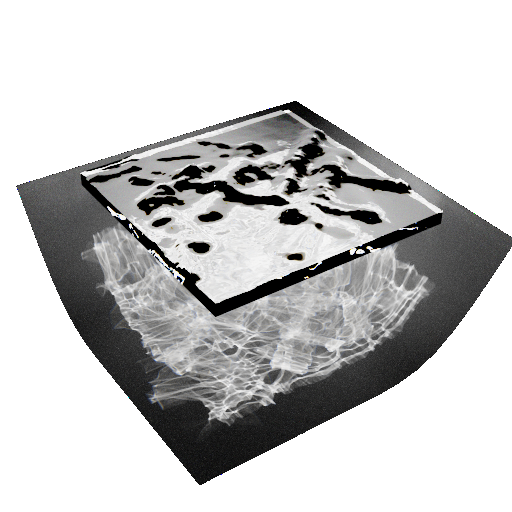}
    \includegraphics[width=0.17\textwidth]{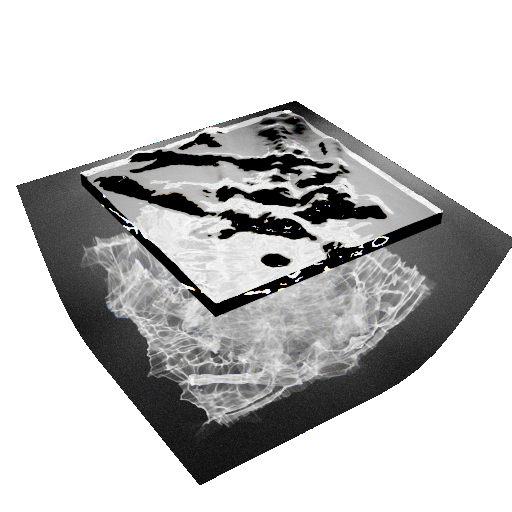}
    \includegraphics[width=0.17\textwidth]{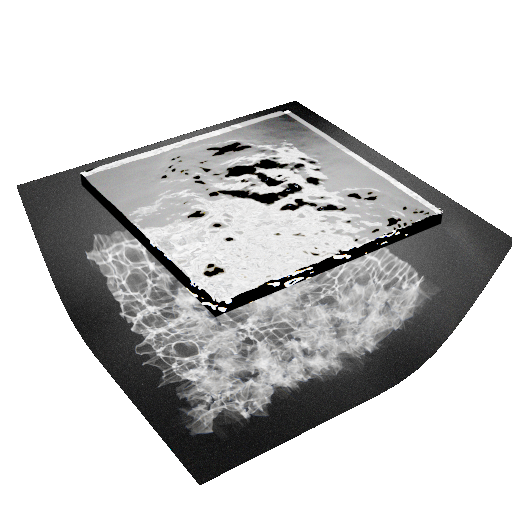}
    \includegraphics[width=0.17\textwidth]{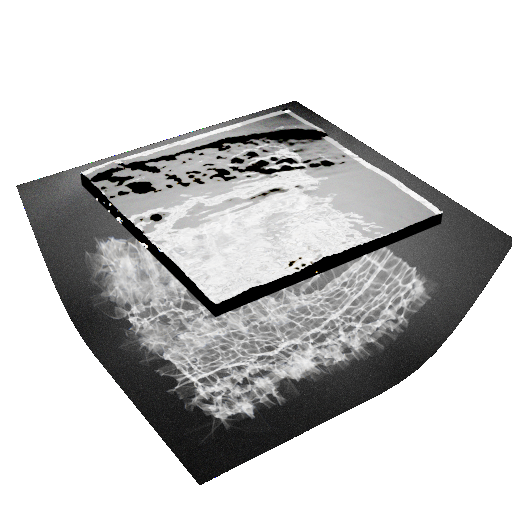}
    \includegraphics[width=0.17\textwidth]{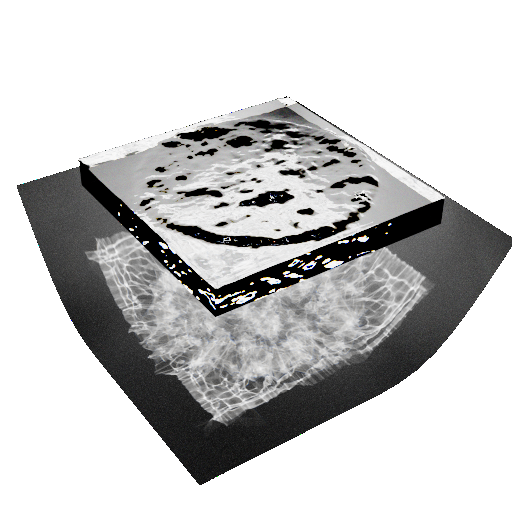}
    \caption{\textbf{Renderings} of N-SfC reconstructions after three update iterations. }
    \label{fig:nsfc_renderings}
\end{figure*}

\begin{figure*}
    \centering
    \includegraphics[width=0.17\textwidth]{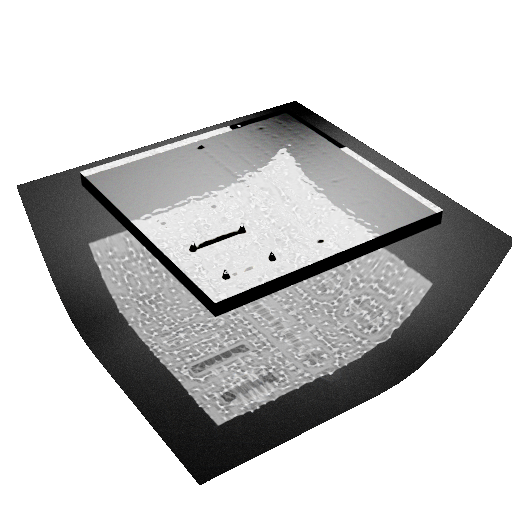}
    \includegraphics[width=0.17\textwidth]{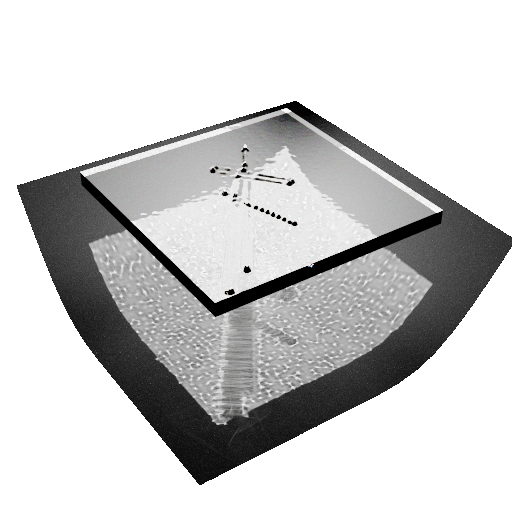}
    \includegraphics[width=0.17\textwidth]{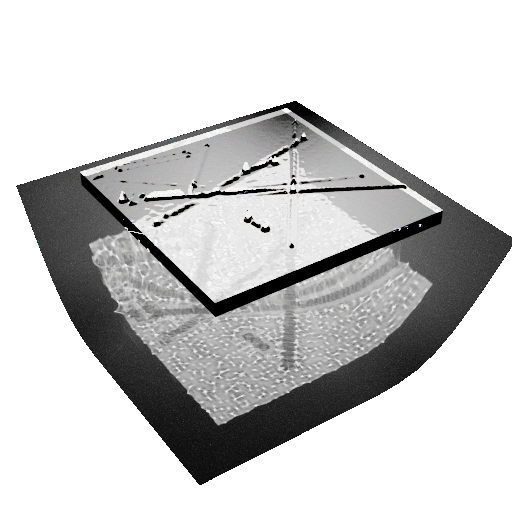}
    \includegraphics[width=0.17\textwidth]{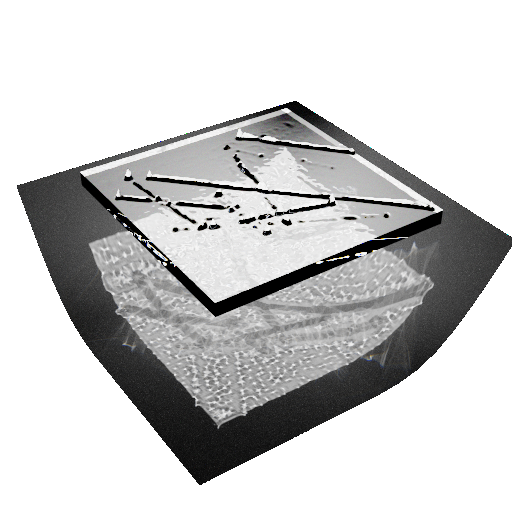}
    \includegraphics[width=0.17\textwidth]{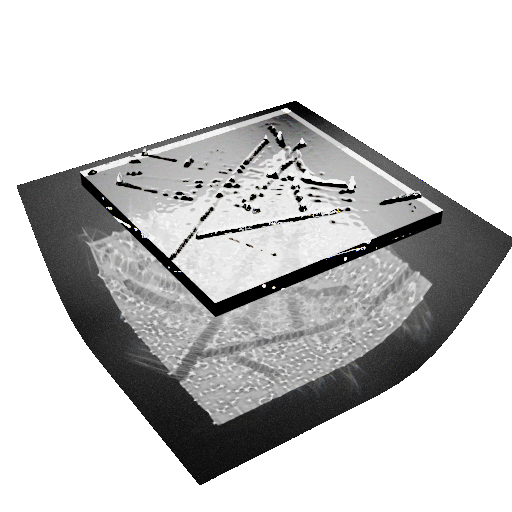}\\
    \includegraphics[width=0.17\textwidth]{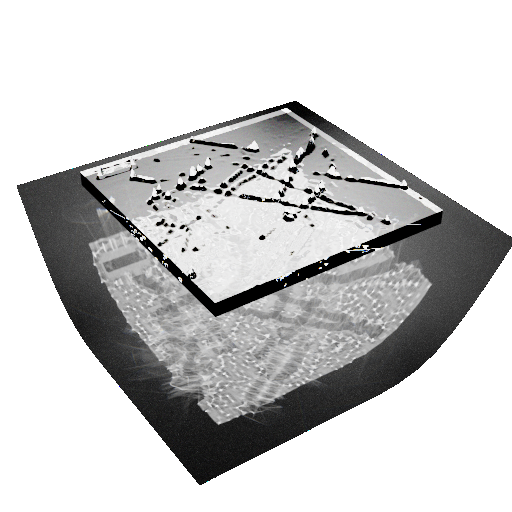}
    \includegraphics[width=0.17\textwidth]{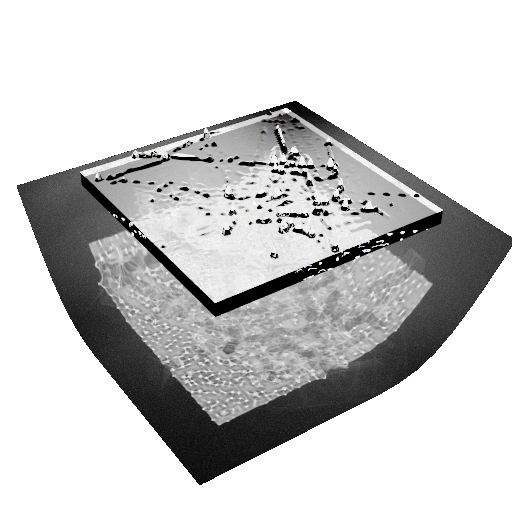}
    \includegraphics[width=0.17\textwidth]{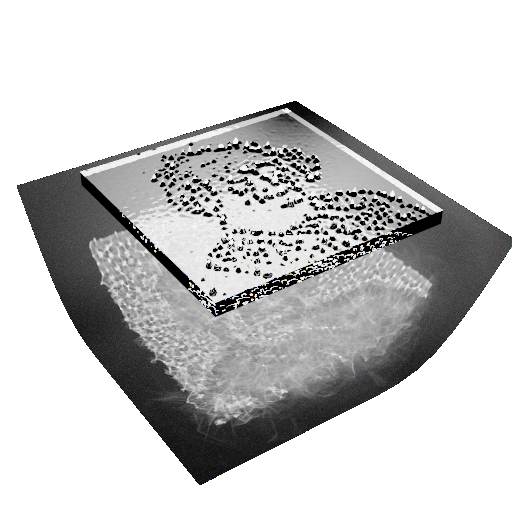}
    \includegraphics[width=0.17\textwidth]{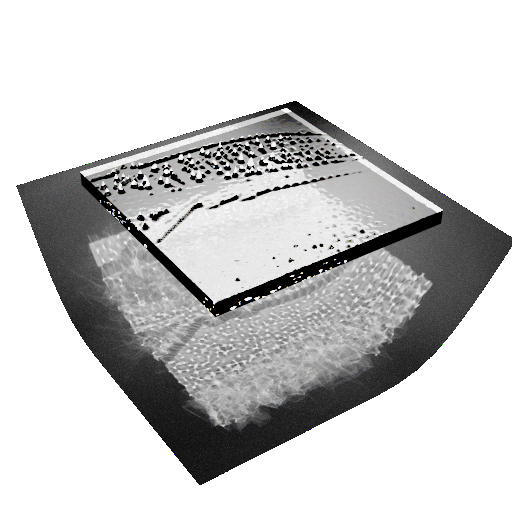}
    \includegraphics[width=0.17\textwidth]{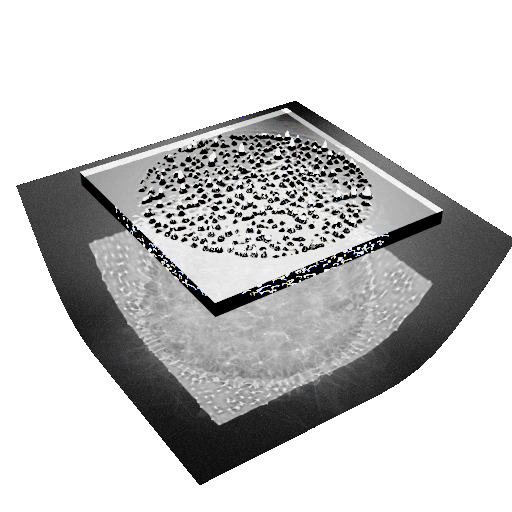}
    \caption{\textbf{Renderings} of SfC reconstructions after one update iteration. }
    \label{fig:sfc_renderings}
\end{figure*}

We provide additional comparison of SfC against N-SfC in \cref{fig:sfc_nsfc_convergence}, where we plotted the relative error $\mathbf{L_rel}$ against the number of update steps taken.
It is visible that SfC is able to decrease the shape error in the first update iteration and sometimes even in the second one as well, but in later iterations the error rises consistently over the error of the initial guess, which is the constant substrate thickness.
N-SfC consistently outperforms the respective errors even in the first iteration and achieves minimal shape error in less than twelve iterations for all samples considered here.
Due to the dependence of the optimal number of update steps on the complexity of the reconstructed height field, we report errors and show results for N-SfC after three update steps and for SfC after one update step, as for our dataset these give the best reconstruction on average.
We provide renderings of the reconstruction results in \cref{fig:nsfc_renderings} and \cref{fig:sfc_renderings}.

\section{Ablation}
\begin{figure*}
    \centering
    \includegraphics[width=0.17\textwidth]{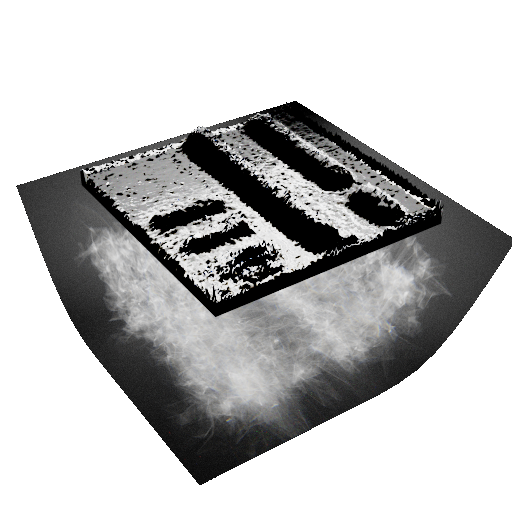}
    \includegraphics[width=0.17\textwidth]{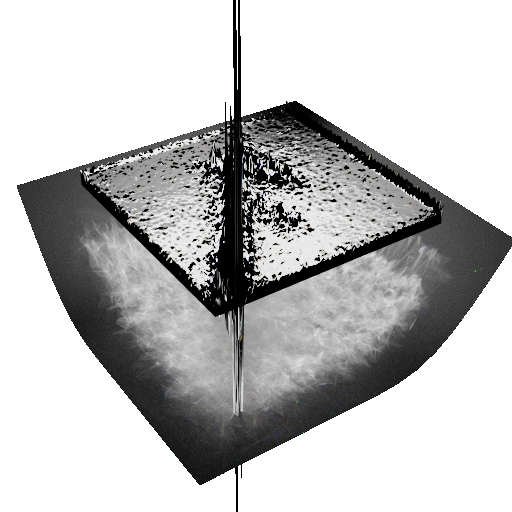}
    \includegraphics[width=0.17\textwidth]{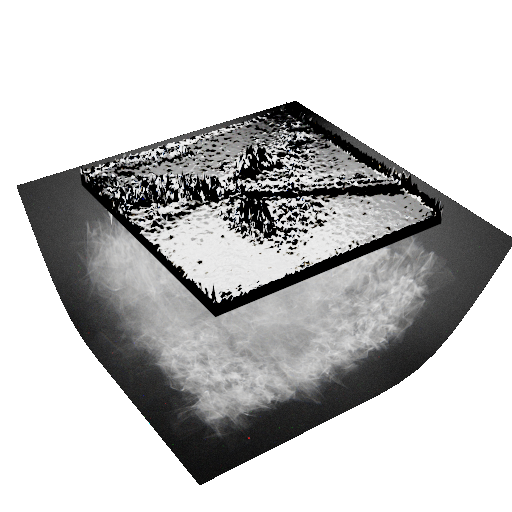}
    \includegraphics[width=0.17\textwidth]{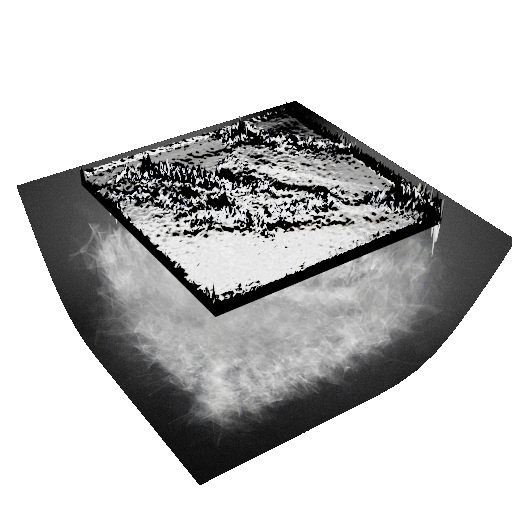}
    \includegraphics[width=0.17\textwidth]{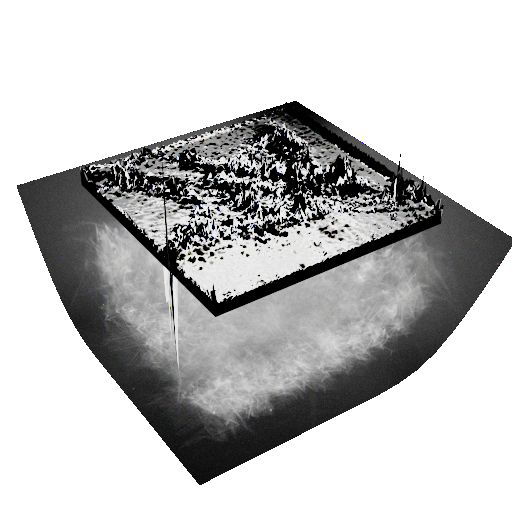}\\
    \includegraphics[width=0.17\textwidth]{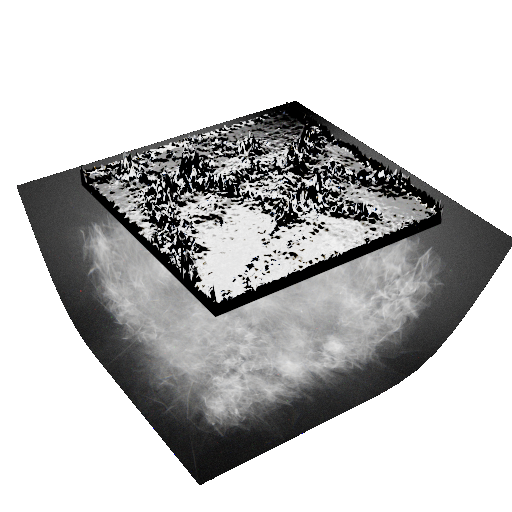}
    \includegraphics[width=0.17\textwidth]{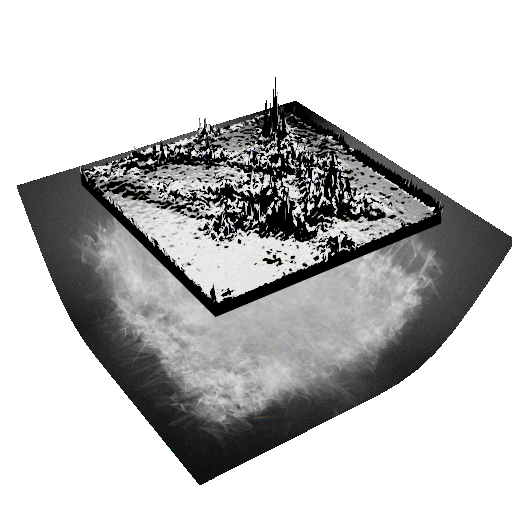}
    \includegraphics[width=0.17\textwidth]{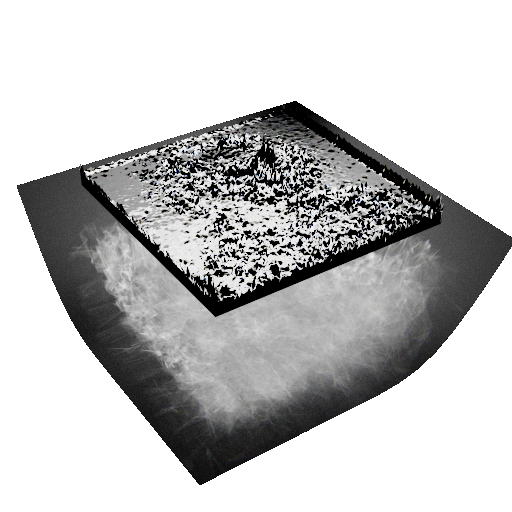}
    \includegraphics[width=0.17\textwidth]{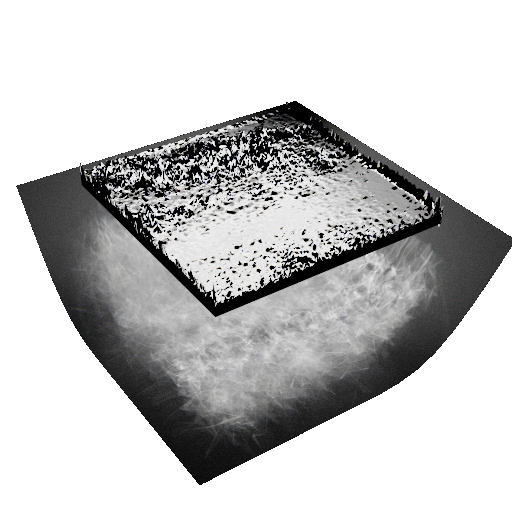}
    \includegraphics[width=0.17\textwidth]{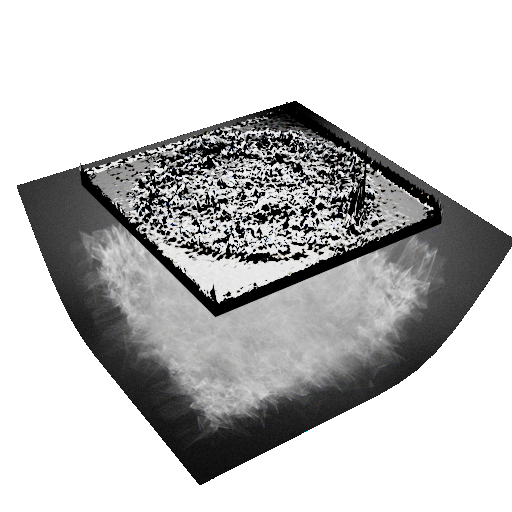}
    \caption{\textbf{Renderings} of N-SfC reconstructions without the denoiser after three update iterations. }
    \label{fig:ablation_denoiser}
\end{figure*}

\begin{figure*}
    \centering
    \includegraphics[width=0.17\textwidth]{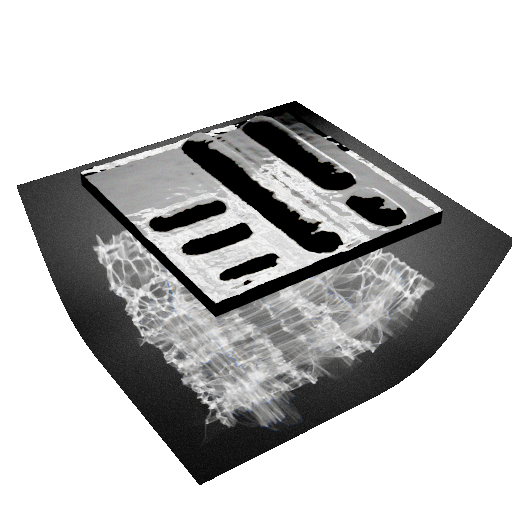}
    \includegraphics[width=0.17\textwidth]{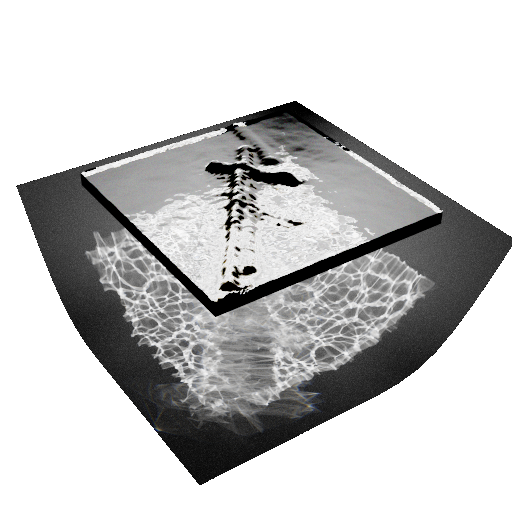}
    \includegraphics[width=0.17\textwidth]{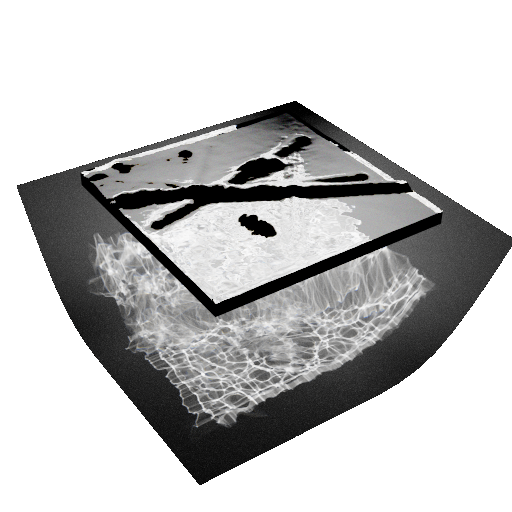}
    \includegraphics[width=0.17\textwidth]{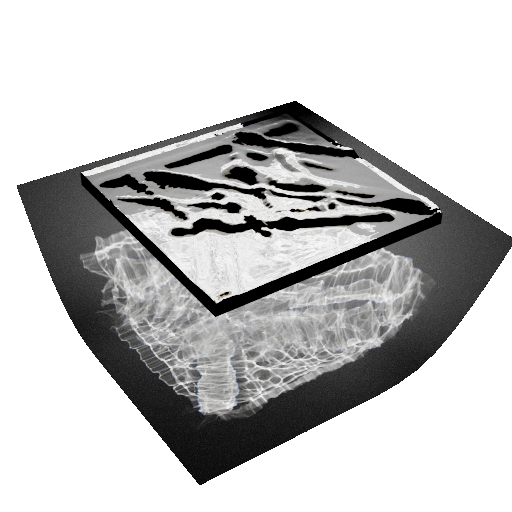}
    \includegraphics[width=0.17\textwidth]{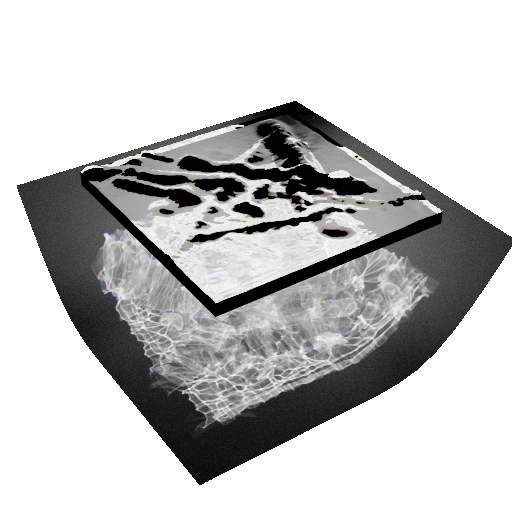}\\
    \includegraphics[width=0.17\textwidth]{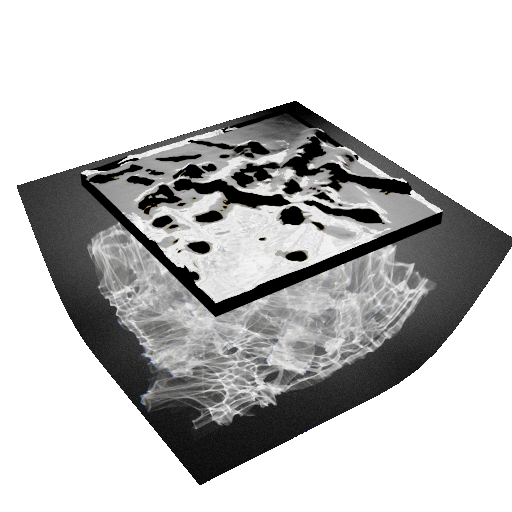}
    \includegraphics[width=0.17\textwidth]{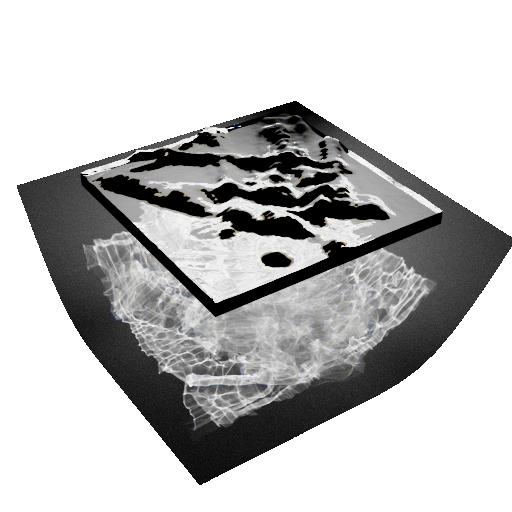}
    \includegraphics[width=0.17\textwidth]{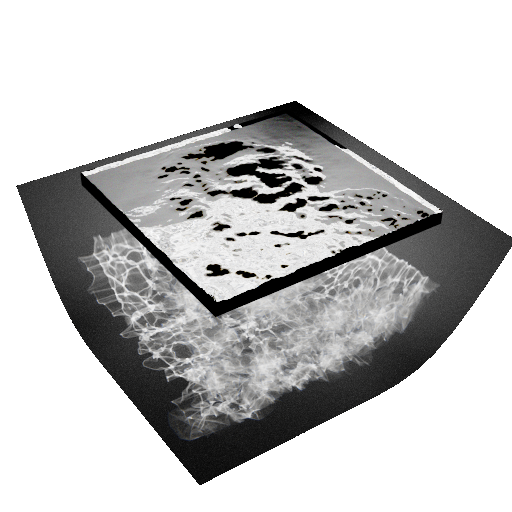}
    \includegraphics[width=0.17\textwidth]{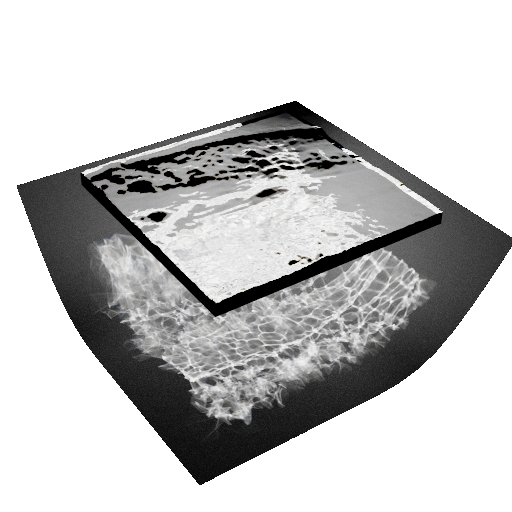}
    \includegraphics[width=0.17\textwidth]{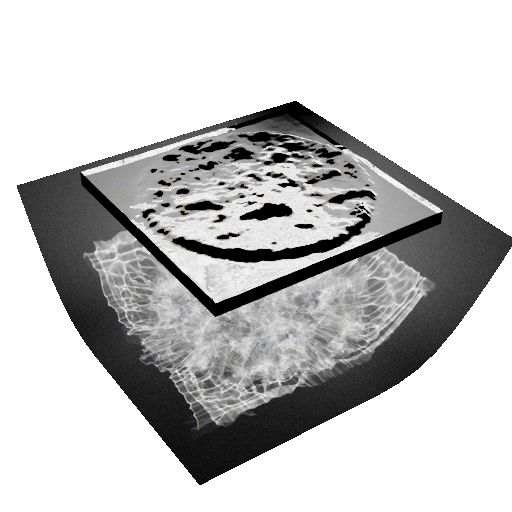}
    \caption{\textbf{Renderings} of N-SfC reconstructions without the local gradient information after three update iterations. }
    \vspace*{9in}
    \label{fig:ablation_gradient}
\end{figure*}

We provide renderings of our ablations, \ie the updater trained without the denoiser in place and the updater without the gradient information from the differentiable renderer in \cref{fig:ablation_denoiser} and \cref{fig:ablation_gradient}.
Face validity of the visualizations shows that the denoiser is an essential part of the framework and reconstruction fails without this component, whereas reasonable reconstructions can still be achieved without the local gradient information.

\end{document}